%% file: sample.tex
\documentclass{applemlr}

\input{apple_preamble}

\title{Unmasking On-Policy Distillation: \\ Where It Helps, Where It Hurts, and Why}

\author[*]{Mohammadreza Armandpour}
\author[*]{Fatih Ilhan}
\author{David Harrison}
\author{Ajay Jaiswal}
\author{Duc N.M Hoang}
\author{Fartash Faghri}
\author{Yizhe Zhang}
\author{Minsik Cho}
\author{Mehrdad Farajtabar}

\affiliation{Apple}
\contribution[*]{Equal contribution}

\abstract{
On-policy distillation offers dense, per-token supervision for training reasoning models; however, it remains unclear under which conditions this signal is beneficial and under which it is detrimental.
Which teacher model should be used, and in the case of self-distillation, which specific context should serve as the supervisory signal?
Does the optimal choice vary from one token to the next?
At present, addressing these questions typically requires costly training runs whose aggregate performance metrics obscure the dynamics at the level of individual tokens.
We introduce a training-free diagnostic framework that operates at the highest resolution: per token, per question, and per teacher.
We derive an ideal per-node gradient defined as the parameter update that maximally increases the student's probability of success. We then develop a scalable targeted-rollout algorithm to estimate this gradient efficiently, even for long chains of intermediate thoughts.
The \textbf{gradient alignment score}, defined as the cosine similarity between this ideal gradient and any given distillation gradient, quantifies the extent to which a particular configuration approximates the ideal signal.  
Across a range of self-distillation settings and external teacher models, we observe that distillation guidance exhibits substantially higher alignment with the ideal on incorrect rollouts than on correct ones, where the student already performs well and the teacher’s signal tends to become noisy. Furthermore, we find that the optimal distillation context depends jointly on the student model’s capacity and the target task, and that no single universally effective configuration emerges. These findings motivate the use of per-task, per-token diagnostic analyses for distillation.
}

\metadata[Correspondence]{\sffamily Mohammadreza Armandpour, Mehrdad Farajtabar \texttt{\{marmandpour, farajtabar\}@apple.com}}
\date{\sffamily\today}

\begin{document}

\maketitle

\section{Introduction}
\label{sec:intro}

On-policy distillation has rapidly become a core post-training technique for reasoning models: Qwen3~\citep{qwen3}, MiMo~\citep{xiao2026mimo}, and GLM-5~\citep{zeng2026glm5} all adopt it in their pipelines, and multiple concurrent works~\citep{hubotter2026sdpo, zhao2026opsd, ye2026onpolicy, shenfeld2026continual} demonstrate strong gains from self-distillation variants, establishing it as a practical and compute-efficient complement to sparse-reward RL.
The idea is simple: guide the student at every token using a teacher's distribution~\citep{agarwal2024gkd, thinking2025distillation}.
In \emph{teacher distillation}, a larger model provides supervision~\citep{hinton2015distilling}.
In \emph{self-distillation}, the student serves as its own teacher with extra context (such as a correct solution) unavailable at test time.
Both complement the sparse binary reward of RL methods like GRPO~\citep{shao2024deepseekmath, deepseek2025r1} with a dense gradient at every token.

Yet practitioners face a series of decisions with no principled guidance: Should the teacher be a larger external model, or the student itself with access to a correct solution?
Should the context include a full solution trace or a concise summary?
Does the answer depend on the question? On the token?
Today, these questions require expensive training runs whose aggregate metrics hide what happens at the level of individual tokens.

\begin{figure}[t]
\centering
\resizebox{\textwidth}{!}{%
\input{figures/figure1}%
}
\caption{\textbf{Not all teacher guidance points toward success.} Generation tree for a bookshelf problem. The teacher's distribution disagrees with the student at four branching points (orange), but not all disagreements are equal: some reflect stylistic preferences (``four'' vs.\ ``4'', ``therefore'' vs.\ ``so,'') rather than reasoning corrections. Standard distillation treats all four signals equally, mixing noise from irrelevant preferences with updates that actually matter.}
\label{fig:bookshelf}
\end{figure}
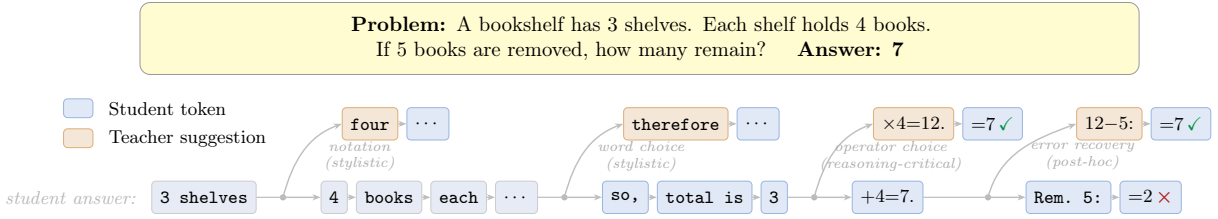

Our objective was to develop a more rigorous methodology: a framework capable of assessing, at the finest feasible level of granularity (\emph{per token, per question, per teacher configuration}), the extent to which the teacher’s guidance is aligned with the behaviors that yield correct answers.
Figure~\ref{fig:bookshelf} demonstrates that, even within an individual reasoning trajectory, the teacher’s points of disagreement comprise a heterogeneous mixture of beneficial, neutral, and detrimental contributions, which cannot be reliably differentiated without explicitly linking each token to its downstream effects.

To evaluate teacher guidance quality at each token, we derive an ideal per-token gradient from empirical success probabilities: the direction that maximally improves the student's chance of reaching a correct answer.
We show that Dr.\ GRPO~\citep{liu2025understanding} recovers this gradient in expectation, making it an unbiased estimator of the ideal (Section~\ref{sec:method-ideal}).
We further show that major distillation objectives (GKD~\citep{agarwal2024gkd}, the single-sample estimator of \citet{thinking2025distillation}, MiniLLM~\citep{gu2024minillm}) produce gradients with the same local structure: for reward-based methods the signal comes from success probability, for distillation methods it comes from the teacher's distribution.
To estimate the ideal gradient scalably even for long reasoning chains, we design a targeted-rollout algorithm with exponential depth windows whose compute scales with a user-chosen budget rather than sequence length.
The \textbf{gradient alignment score} (cosine similarity between the ideal and the distillation gradient at each token) then evaluates how well any teacher configuration approximates the ideal, offline (Section~\ref{sec:method-score}).

\paragraph{Key findings.}
Applying this framework to Qwen3-0.6B and Qwen3-1.7B across 8 teacher configurations on BoolQ and MMLU, we find that:
\begin{itemize}[leftmargin=*, itemsep=1pt]
  \item \textbf{Distillation guidance is more reliable on incorrect rollouts.} When the student is already on a correct path, the teacher's signal becomes noisy and weakly aligned with the ideal; on failing rollouts, the teacher reliably pushes toward success. This holds across all settings and metrics.
  \item \textbf{Context design and student capacity interact strongly.} In self-distillation, the form of context shown to the student-as-teacher matters: a summarized solution nearly doubles alignment for 1.7B compared to the raw trace, but slightly hurts 0.6B (which needs full step-by-step reasoning). A 32B-generated solution helps 0.6B on simple tasks but fails on hard math where the reasoning style becomes foreign. External teachers outperform self-distillation only for the larger student. We hypothesize that \emph{comprehensibility} is the underlying factor: the gradient signal is only useful if the student can parse what it is given.
  \item \textbf{No universal recipe exists.} Among self-distillation variants, contrastive examples (correct + wrong) hurt on simple reasoning but help on hard math. Comparing self-distillation to external teachers, external teachers outperform for 1.7B on BoolQ but not on MMLU. Which teacher or context achieves the highest alignment shifts with question difficulty, motivating per-task diagnostics rather than fixed pipelines.
  \item \textbf{Divergence predicts alignment, but weakly.} Within-path correlations show that divergence between student and teacher prediction distributions (KL, JS, $L_2$) is positively associated with alignment while their similarity (cosine of probability vectors) is negatively associated, consistently across all settings. Magnitudes are small ($|\rho| < 0.05$), indicating divergence as a cheap necessary-condition filter but not a reliable predictor.
\end{itemize}
We further test these patterns on AIME 2025  as case studies (Section~\ref{sec:aime}); the finding that incorrect rollouts exhibit higher alignment replicates, but the best self-distillation context changes: including a wrong demonstration, which hurts on short-reasoning tasks, produces the highest alignment on hard math problems.

\input{method}

\section{Experimental Setup}
\label{sec:setup}

\paragraph{Student models.}
We evaluate two student scales from the Qwen3 family~\citep{qwen3}: Qwen3-0.6B and Qwen3-1.7B.

\paragraph{Teacher configurations.}
For each student, we evaluate 8 teacher configurations spanning two families:

\emph{Self-distillation} (teacher = same model with enriched context):
\textsc{Self-1C} (1 correct solution in context),
\textsc{Self-Sum-1C} (correct solution summarized by Qwen3-32B),
\textsc{Self-1C1W} (1 correct + 1 wrong solution),
\textsc{Self-Sum-1C1W} (both summarized),
\textsc{Self-1C\,(32B)} (correct solution generated by Qwen3-32B shown to student-as-teacher).

\emph{External teachers} (larger models, same prompt as student):
Qwen3-4B, Qwen3-8B, Qwen3-14B.

\paragraph{Datasets.}
We evaluate on two benchmarks:
\textbf{BoolQ}~\citep{clark2019boolq}, a reading comprehension task with True/False answers and short reasoning chains; and
\textbf{MMLU}~\citep{hendrycks2021mmlu}, a multiple-choice knowledge benchmark with medium-length chains.
We additionally present case studies on AIME 2025 ($\sim$5K--30K token traces) in Section~\ref{sec:aime}.
Each question requires substantial compute: $G{=}200$ initial rollouts, 4 representative paths (2 correct, 2 incorrect), and $\sim$45K--200K targeted rollouts at undersampled branching points (totaling $\sim$72 A100-days for the full experiment suite).
Each important token receives up to 100 targeted samples; nodes with $\geq$20 visits are considered statistically significant for computing $\hat{P}_{\mathrm{succ}}$.

\paragraph{Metrics.}
At each branching node with $\geq$2 children having $\geq$20 visits and nonzero success-rate range, we compute: gradient alignment (ideal vs.\ GKD cosine), teacher advantage, and success rate statistics.
We aggregate per path (mean cosine along the path), per question (correct/incorrect split), and per teacher (means with 95\% CIs across questions).

\section{Results}
\label{sec:experiments}

We present results across two datasets and two model scales, totaling $\sim$88K decision points for BoolQ (0.6B) and $\sim$81K for BoolQ (1.7B), with MMLU providing $\sim$49K (0.6B) and $\sim$46K (1.7B).
Overall, gradient alignment is weakly positive (mean cosine $+0.027$ for 0.6B, $+0.026$ for 1.7B on BoolQ) but with enormous per-token variance (std $\sim$0.83--0.91; see Appendix~\ref{app:histogram} for the full distribution).

\subsection{Distillation helps more on incorrect paths}
\label{sec:correct-incorrect}

\begin{figure}[t]
\centering
\begin{minipage}[c]{0.56\textwidth}
\centering
\includegraphics[width=\linewidth]{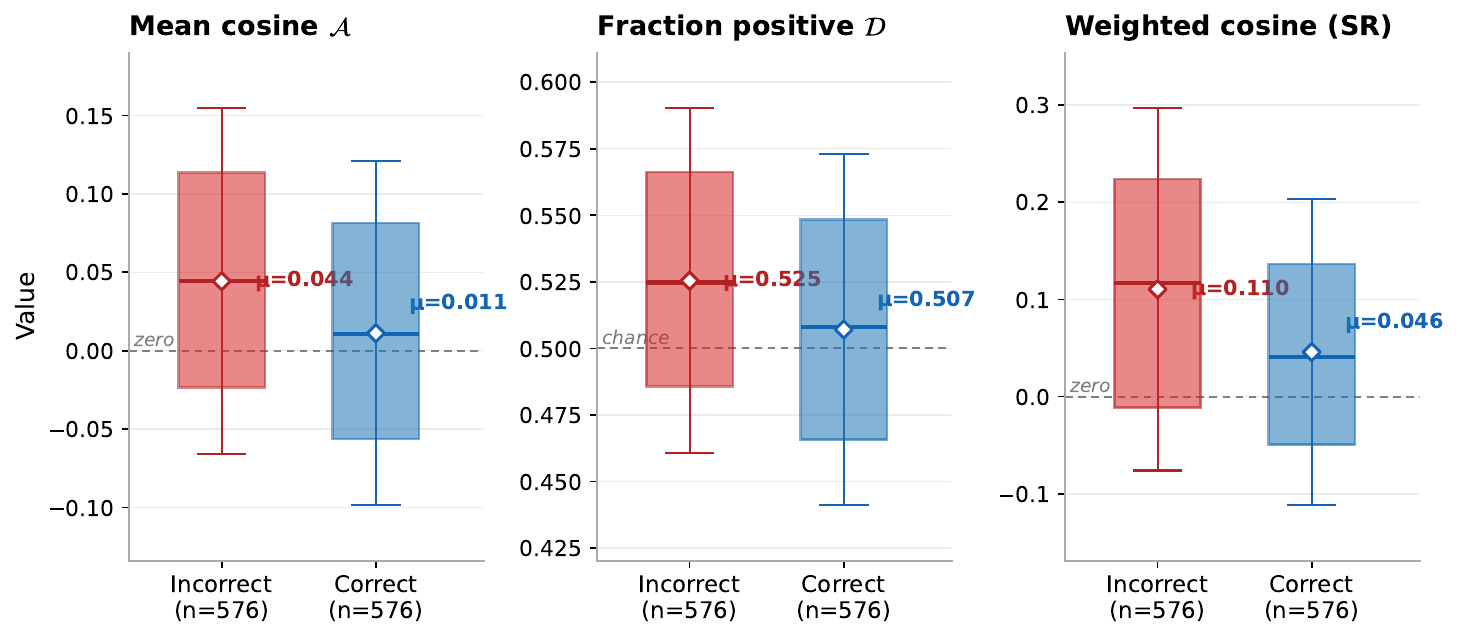}
\end{minipage}%
\hfill
\begin{minipage}[c]{0.42\textwidth}
\centering
\scriptsize
\setlength{\tabcolsep}{3pt}
\begin{tabular}{lccc}
\toprule
& \textbf{Correct} & \textbf{Incorrect} & $p$-value \\
\midrule
\multicolumn{4}{l}{\textit{Qwen3-0.6B, BoolQ}} \\
Mean cosine        & .011 & .044 & $7\!\times\!10^{-8}$ \\
Weighted (SR)      & .046 & .110 & $8\!\times\!10^{-10}$ \\
\midrule
\multicolumn{4}{l}{\textit{Qwen3-1.7B, BoolQ}} \\
Mean cosine        & .001 & .058 & $2\!\times\!10^{-9}$ \\
Weighted (SR)      & .010 & .093 & $7\!\times\!10^{-9}$ \\
\midrule
\multicolumn{4}{l}{\textit{Qwen3-0.6B, MMLU}} \\
Mean cosine        & .009 & .048 & $.0001$ \\
Weighted (SR)      & .021 & .118 & $<\!10^{-11}$ \\
\midrule
\multicolumn{4}{l}{\textit{Qwen3-1.7B, MMLU}} \\
Mean cosine        & .012 & .028 & $.123$ \\
Weighted (SR)      & .034 & .098 & $<\!10^{-5}$ \\
\bottomrule
\end{tabular}
\end{minipage}
\caption{Gradient alignment on correct vs.\ incorrect paths. \textbf{Left:} distribution for Qwen3-0.6B on BoolQ; the teacher's gradient is more aligned on paths leading to wrong answers. \textbf{Right:} the pattern holds across all settings under both mean cosine and SR-weighted cosine. Full results in Appendix Table~\ref{tab:correct-incorrect-full}.}
\label{fig:correct-incorrect}
\end{figure}

Our most consistent finding across all settings is that \textbf{incorrect paths exhibit significantly higher gradient alignment than correct paths} (Figure~\ref{fig:correct-incorrect}).
On incorrect paths, the reward gradient points away from the current (failing) trajectory, and the teacher (which generally prefers tokens leading to success) pushes in the same direction.
On correct paths, the student is already succeeding, so the reward gradient is weaker and the teacher's contribution is less aligned.
The effect is strongest for 1.7B on BoolQ ($\Delta = -0.056$, $p < 10^{-9}$); even on MMLU where the mean cosine gap is not significant ($p=0.12$), the weighted cosine is highly significant ($p < 10^{-5}$).

\subsection{The best teacher depends on student capacity}
\label{sec:teacher-ranking}

\begin{figure}[t]
\centering
\includegraphics[width=\textwidth]{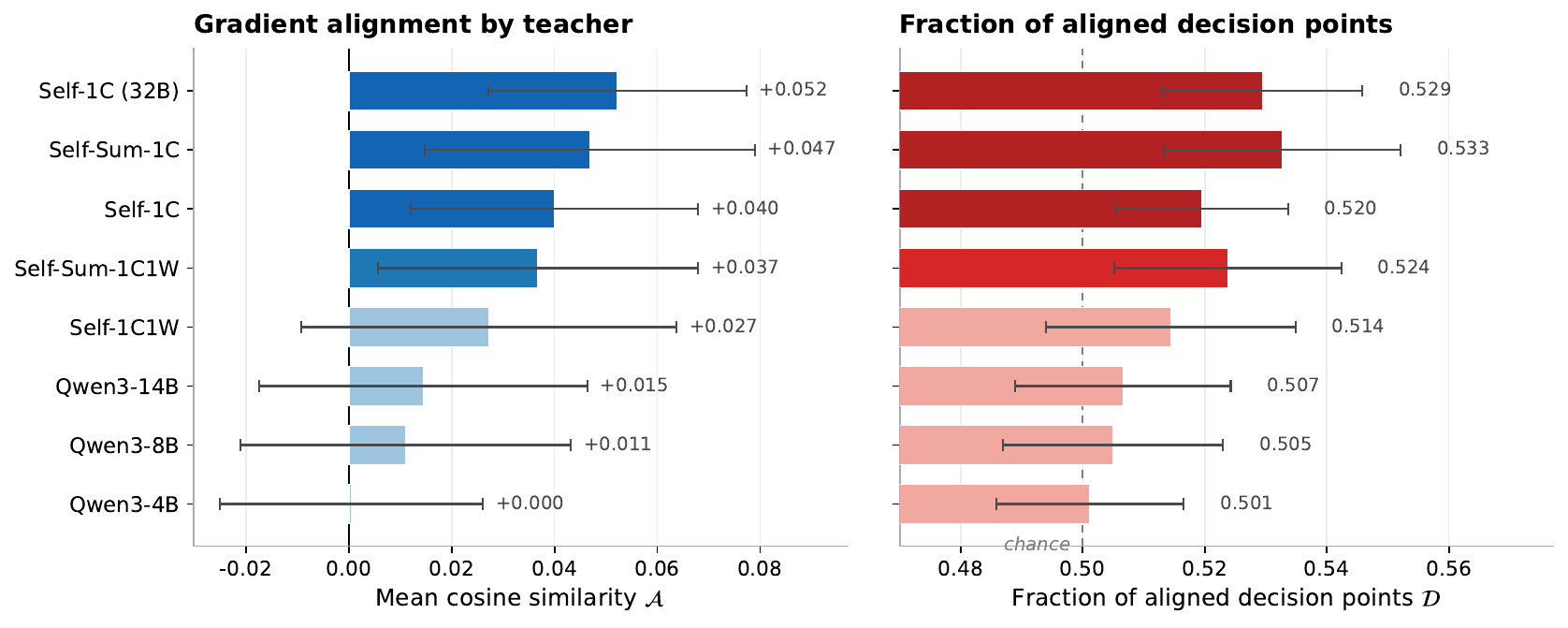}
\caption{Teacher ranking by gradient alignment (Qwen3-0.6B, MMLU). Self-distillation with correct demonstrations dominates. Additional plots for BoolQ and 1.7B in Figure~\ref{fig:teacher-bars-appendix}.}
\label{fig:teacher-bars}
\end{figure}

A striking result emerges when comparing teacher rankings across model scales (Figure~\ref{fig:teacher-bars}, Table~\ref{tab:teacher-ranking}).
For the \textbf{0.6B student}, self-distillation teachers using correct-only demonstrations (\textsc{Self-1C}, \textsc{Self-Sum-1C}, \textsc{Self-1C\,(32B)}) consistently achieve $2$--$3\times$ higher alignment than external teachers, on both BoolQ and MMLU.
But for the \textbf{1.7B student on BoolQ}, external teachers, particularly Qwen3-8B, achieve the highest alignment, outperforming all self-distillation variants.

\begin{table}[t]
\centering
\caption{Teacher ranking by mean gradient alignment (with 95\% CI) across datasets. Self-distillation dominates for 0.6B; external teachers become competitive for 1.7B.}
\label{tab:teacher-ranking}
\small
\begin{tabular}{lcccc}
\toprule
 & \multicolumn{2}{c}{\textbf{Qwen3-0.6B}} & \multicolumn{2}{c}{\textbf{Qwen3-1.7B}} \\
\cmidrule(lr){2-3} \cmidrule(lr){4-5}
\textbf{Teacher} & BoolQ & MMLU & BoolQ & MMLU \\
\midrule
\textsc{Self-1C}            & $\mathbf{.047 \pm .021}$ & $.040 \pm .028$ & $.028 \pm .024$ & $.010 \pm .024$ \\
\textsc{Self-1C\,(32B)}    & $.047 \pm .022$ & $\mathbf{.052 \pm .025}$ & $.020 \pm .034$ & $.021 \pm .030$ \\
\textsc{Self-Sum-1C}       & $.041 \pm .025$ & $.047 \pm .032$ & $\mathbf{.050 \pm .030}$ & $\mathbf{.036 \pm .025}$ \\
\textsc{Self-Sum-1C1W}     & $.016 \pm .025$ & $.037 \pm .031$ & $.008 \pm .029$ & $.034 \pm .033$ \\
\textsc{Self-1C1W}         & $.019 \pm .018$ & $.027 \pm .037$ & $.002 \pm .027$ & $.009 \pm .022$ \\
\midrule
Qwen3-14B       & $.016 \pm .021$ & $.015 \pm .032$ & $.036 \pm .035$ & $.017 \pm .035$ \\
Qwen3-8B        & $.018 \pm .022$ & $.011 \pm .032$ & $\mathbf{.053 \pm .028}$ & $.014 \pm .037$ \\
Qwen3-4B        & $.018 \pm .024$ & $.000 \pm .026$ & $.040 \pm .024$ & $.017 \pm .041$ \\
\bottomrule
\end{tabular}
\end{table}

We interpret these findings through the lens of \textbf{context comprehensibility}: self-distillation helps only when the student can understand the context it is given.
A small student (0.6B) cannot effectively absorb the full distributional knowledge from a much larger external model; the teacher's reasoning patterns are too different from its own, making the gradient signal incomprehensible.
Showing the same 0.6B model a correct solution in its own reasoning style (self-distillation) provides a targeted, understandable nudge toward success tokens.
For the larger 1.7B student, the capacity gap to an 8B teacher is smaller, and the genuinely different knowledge encoded in the larger model's distribution becomes comprehensible and exploitable.

An additional finding: \textbf{including wrong demonstrations hurts} on BoolQ and MMLU. The 1C1W variants consistently underperform their 1C counterparts, suggesting that negative examples introduce noise rather than useful contrastive signal on these tasks.
Breaking down the correct-vs-incorrect gap per teacher (Appendix~\ref{app:per-teacher-delta}), we find that \textsc{Self-1C} for 0.6B is uniquely uniform: it achieves nearly equal alignment on both correct and incorrect paths ($\Delta \approx 0$), while all other teachers show the typical incorrect $>$ correct pattern.

\paragraph{Summarization helps larger students.}
Summarizing demonstrations (\textsc{Self-Sum-1C}) nearly doubles alignment for the 1.7B student ($0.050$ vs.\ $0.028$ on BoolQ, $0.036$ vs.\ $0.010$ on MMLU), but has mixed effects for 0.6B.
The interpretation is capacity-dependent: a larger student can extract the key signal from a concise summary, while a smaller student needs the full verbose trace.
Similarly, \textsc{Self-1C\,(32B)} (a Qwen3-32B-generated solution shown to the student-as-teacher) works well for 0.6B ($0.047$, $0.052$) but poorly for 1.7B and on AIME, as the 32B reasoning style is harder to follow on complex problems.
These rankings are largely robust to metric choice (Appendix~\ref{app:teacher-alt}), though for 1.7B the weighted cosine reverses the top ranking (\textsc{Self-Sum-1C}: $0.088$ vs.\ Qwen3-8B: $0.072$), suggesting the external teacher's advantage is concentrated at low-stakes tokens.

\subsection{What predicts alignment within a path?}
\label{sec:correlations}

To understand \emph{where} alignment is positive within a reasoning chain, we compute within-path Spearman correlations between the alignment score and single-rollout features (full results in Appendix~\ref{app:correlations}).
Teacher--student divergence (KL, JS, $L_2$) correlates positively with alignment, while distributional similarity correlates negatively: the useful signal lives where the teacher disagrees with the student.
However, correlations are uniformly weak ($|\rho| \approx 0.02$--$0.04$): no single feature reliably predicts whether a disagreement is helpful or harmful, since that depends on downstream success probabilities not observable from a single forward pass.
The consistent sign nonetheless suggests divergence could serve as a cheap necessary-condition filter for alignment-aware training.
Alignment also trends slightly positive with depth (early tokens are templatic; later tokens involve actual reasoning steps).

\subsection{Case studies: mathematical reasoning (AIME 2025)}
\label{sec:aime}

To test generalization to longer reasoning traces ($\sim$5K--30K tokens), we analyze 4 AIME 2025 questions (2 per model) using the same 8 teacher configurations (full results in Appendix~\ref{app:aime}).
The core finding (incorrect paths exhibit higher alignment) replicates across all four questions (e.g., best teacher on Q0: incorrect $+0.097$ vs.\ correct $-0.011$, $\Delta = -0.108$).
However, teacher-choice conclusions diverge from BoolQ/MMLU: \textsc{Self-1C1W} (including a wrong demonstration) is the \emph{best} teacher on the two harder questions, directly contradicting the shorter-reasoning benchmarks where wrong demos consistently hurt.
The interpretation is comprehensibility-dependent: on hard math, seeing a common mistake provides useful contrastive signal, whereas on simpler tasks it is merely noise.
Additionally, summarized contexts lose to raw demos on hard math (the 0.6B student cannot decipher a compressed summary of a complex argument) but perform well on easier questions.
These observations reinforce that no universal distillation recipe exists: the optimal teacher depends on task difficulty, student capacity, and context design.

\input{related_work_short}

\section{Conclusion}
\label{sec:conclusion}

We set out to answer a simple question: at each token in a reasoning chain, does the teacher's distillation signal actually point toward correct answers?
We derived an ideal per-node gradient from empirical success probabilities, showed that major distillation objectives share the same local structure, and built a scalable pipeline to compute gradient alignment offline for long sequences.

Our experiments reveal that distillation helps most on failing rollouts, that context design interacts strongly with student capacity, and that no universal recipe exists.
Beyond these findings, the framework serves as a general-purpose offline testbed for any token-level training algorithm.
Our results point to concrete future directions: rollout-weighted distillation that emphasizes failing trajectories, multi-teacher distillation that combines complementary signals from multiple teachers for better overall alignment, adaptive context selection per domain, and divergence-based filters for alignment-aware training.

\bibliographystyle{plainnat}
\bibliography{references}

\appendix

\section{Full AIME 2025 Case Studies}
\label{app:aime}

\input{aime_results}

\section{Full Gradient Derivations}
\label{app:derivations}

\subsection{Softmax Jacobian}

The student's transition probability at node $u$ is $P_\theta^k = e^{z_k} / \sum_{k'} e^{z_{k'}}$, with Jacobian:
\begin{equation}
\frac{\partial P_\theta^k}{\partial z_j} = P_\theta^k(\delta_{kj} - P_\theta^j)
\end{equation}

For any objective $L(u) = \sum_k P_\theta^k f_k$ where $f_k$ does not depend on $\theta$:
\begin{equation}
\frac{\partial L}{\partial z_j} = \sum_k f_k \cdot P_\theta^k(\delta_{kj} - P_\theta^j) = P_\theta^j f_j - P_\theta^j \sum_k P_\theta^k f_k = P_\theta^j(f_j - \bar{f})
\end{equation}

\subsection{Dr.\ GRPO}

The full GRPO objective in the limit $G \to \infty$ is:
\begin{equation}
J_{\mathrm{GRPO}} = \mathbb{E}_{o_i \sim \pi_{\theta_{\text{old}}}} \left( \frac{1}{|o_i|} \sum_t \frac{\pi_\theta(o_{i,t} \mid q,\, o_{i,<t})}{\pi_{\theta_{\text{old}}}(o_{i,t} \mid q,\, o_{i,<t})} \cdot A_i - \beta\, D_{KL}\!\left(\pi_\theta \| \pi_{\text{ref}} \right) \right)
\end{equation}

We make the following simplifications to obtain a per-node decomposition:

\paragraph{1. Drop the KL penalty.} Since $\beta$ is typically small, we set $\beta = 0$.

\paragraph{2. Marginalize the importance ratio.} For trajectories passing through node $u$ and choosing token $k$, the importance weight $\pi_\theta(o_{i,t} \mid \cdot) / \pi_{\theta_\mathrm{old}}(o_{i,t} \mid \cdot) = P_\theta^k / P_{\theta_\mathrm{old}}^k$ at step $t$ cancels with the sampling measure $P_{\theta_\mathrm{old}}^k$ after marginalization, leaving a net factor of $P_\theta^k$.

\paragraph{3. Remove length normalization (Dr.\ GRPO).} Standard GRPO divides by $|o_i|$, coupling the advantage to trajectory length:
\begin{equation}
J(u) \propto \sum_k P_\theta^k \cdot \mathbb{E}\!\left[\frac{A_i}{|o_i|} \;\middle|\; o_i \text{ passes } u \to v_k\right]
\end{equation}
Since $A_i$ and $|o_i|$ are correlated (e.g., successful trajectories may be shorter), this expectation does not factor.
Dr.\ GRPO removes the $1/|o_i|$ factor, giving:
\begin{equation}
J_{\mathrm{DrGRPO}}(u) \propto \sum_k P_\theta^k \cdot \mathbb{E}\!\left[A_i \;\middle|\; o_i \text{ passes } u \to v_k\right]
\end{equation}

\paragraph{4. Evaluate the conditional expectation.} With $A_i = (R_i - \bar{R})/\mathrm{std}(R)$ and binary rewards $R_i \in \{0,1\}$:
\begin{equation}
\mathbb{E}[A_i \mid o_i \text{ passes } u \to v_k] = \frac{P_{\mathrm{succ}}^k - \bar{R}}{\mathrm{std}(R)} = \alpha \cdot P_{\mathrm{succ}}^k - C
\end{equation}
where $\alpha = 1/\mathrm{std}(R)$ and $C = \bar{R}/\mathrm{std}(R)$ are batch-level constants (each trajectory shifts them by $\pm 1/G$, negligible for large $G$).

\paragraph{5. Remove the constant.} Since $\sum_k P_\theta^k \cdot (-C) = -C$ is independent of $\theta$, the gradient-relevant objective at node $u$ reduces to:
\begin{equation}
L_{\mathrm{DrGRPO}}(u) = \sum_k P_\theta^k \cdot P_{\mathrm{succ}}^k
\end{equation}
which is the ideal objective (Equation~\ref{eq:ideal}), giving gradient $P_\theta^j(P_{\mathrm{succ}}^j - \bar{P}_{\mathrm{succ}})$.

\paragraph{Empirical estimator.} In practice with finite samples, the per-node gradient is estimated as:
\begin{equation}
\frac{\partial \hat{L}_{\mathrm{DrGRPO}}}{\partial z_j}\bigg|_u
= \frac{1}{N_u}\sum_{i:\, o_i \ni u} A_i \left(\delta_{r_i, j} - P_\theta^j\right)
\end{equation}
where $r_i$ is the token chosen by rollout $i$ at node $u$ and $N_u$ is the number of rollouts passing through $u$.
In our framework, we instead compute the ideal gradient directly from empirical $\hat{P}_{\mathrm{succ}}^k$ estimates, which is equivalent in expectation but lower variance.

\subsection{GKD}

The GKD loss is $L_{\mathrm{GKD}} = \mathrm{KL}(\pi_\theta \| \pi_{\mathrm{te}}) = \sum_k P_\theta^k (\log P_\theta^k - \log P_{\mathrm{te}}^k)$.
Defining $\ell_k = \log P_\theta^k - \log P_{\mathrm{te}}^k$, the gradient is:
\begin{equation}
\frac{\partial L_{\mathrm{GKD}}}{\partial z_j} = P_\theta^j(\ell_j - \bar{\ell}) \quad \text{where} \quad \bar{\ell} = \mathrm{KL}(\pi_\theta \| \pi_{\mathrm{te}})
\end{equation}
The additional term from $\partial \log P_\theta^k / \partial z_j$ produces $+P_\theta^j(1 - 1) = 0$ after applying the Jacobian sum-to-zero property.

\subsection{Single-sample GKD estimator (Thinking-Lab)}

The empirical estimator uses importance weighting at the sampled token only.
For rollout $i$ choosing token $r_i$ at node $u$, the weight is:
\begin{equation}
  w_i = \log P_{\mathrm{te}}^{r_i} - \log P_\theta^{r_i} - 1
\end{equation}
and the per-sample gradient contribution is:
\begin{equation}
  \nabla_{z_j} \ell_i = w_i \cdot (\delta_{r_i, j} - P_\theta^j)
\end{equation}
Taking the expectation over the student's sampling distribution:
\begin{align}
  \mathbb{E}[\nabla_{z_j} \ell]
  &= \sum_k P_\theta^k \left(\log P_{\mathrm{te}}^k - \log P_\theta^k - 1\right)(\delta_{kj} - P_\theta^j) \nonumber \\
  &= P_\theta^j(-\ell_j - 1) - P_\theta^j \sum_k P_\theta^k(-\ell_k - 1) \nonumber \\
  &= -P_\theta^j(\ell_j - \bar{\ell})
\end{align}
The constant $-1$ cancels because $\sum_k P_\theta^k (\delta_{kj} - P_\theta^j)(-1) = 0$.
This recovers the GKD gradient (Equation~\ref{eq:gkd-grad}) with opposite sign, confirming that the single-sample importance-weighted estimator and the full-vocabulary KL minimization produce the same gradient direction in expectation.

\subsection{MiniLLM}

MiniLLM~\citep{gu2024minillm} uses a REINFORCE-style gradient where the reward-to-go couples each node's gradient to all downstream nodes.
At node $u$ (step $t$, sampled token $r_i$), the per-sample gradient is:
\begin{equation}
  \nabla_{z_j} \ell_i = -(\delta_{r_i, j} - P_\theta^j) \cdot \left(\sum_{t' \geq t} R_{t'} - 1\right)
\end{equation}
where $R_{t'} = \log P_{\mathrm{te}}^{o_{i,t'}} - \log P_\theta^{o_{i,t'}}$ is the per-step reward.
Unlike GKD and its single-sample variant, here $f_k$ in the unified form $P_\theta^j(f_j - \bar{f})$ depends on the full future trajectory rather than being purely local, making the per-node contribution path-dependent.

\section{Discussion, Limitations, and Future Work}
\label{app:discussion}

\paragraph{Limitations.}
The alignment score is restricted to the set of tokens $\mathcal{S}_u$ with sufficient visit counts at each node; tokens that are never sampled by the student cannot be evaluated.
Additionally, targeted rollout enrichment requires substantial compute per question, making the diagnostic most practical as an offline evaluation tool rather than a real-time training signal.

\paragraph{Future directions.}
Key extensions include: (i)~designing training algorithms that exploit our observations, for example up-weighting the distillation loss on incorrect rollouts where alignment is strongest, or gating the teacher signal by student-teacher divergence (which correlates positively with alignment); (ii)~multi-teacher distillation that combines complementary signals from multiple teachers for better overall alignment; (iii)~extending to long-horizon agentic tasks, where per-question variability and context-dependent teacher choice are likely even more pronounced.

\section{Computation Details}
\label{app:computation}

\paragraph{Rollout prioritization.}
Not all nodes are equally informative for targeted enrichment.
We prioritize using two criteria: (i)~GKD gradient magnitude $|P_\theta^j(\ell_j - \bar\ell)|$, identifying tokens where the teacher disagrees most strongly; and (ii)~probability difference $|P_{\mathrm{te}}^j - P_\theta^j|$, identifying where the two distributions diverge most.
We allocate a budget per depth window (smaller, denser windows early; larger, coarser windows later) and rank candidate tokens by both criteria.

\paragraph{Multi-teacher pipeline.}
Phase~1 (shared): generate $G$ initial rollouts, build the generation tree, and run targeted rollouts to enrich branching points. Each teacher's logits are computed to identify important tokens; rollouts accumulate in the shared tree so later teachers benefit from prior enrichment.
Phase~2 (per-teacher): compute $P_{\mathrm{succ}}^k$ once from the enriched tree (teacher-independent), then for each teacher run one forward pass to obtain $P_{\mathrm{te}}^k$ and evaluate the alignment score.

\section{Additional Figures and Analysis}
\label{app:additional-figures}

\subsection{Alignment score distribution}
\label{app:histogram}

The distribution of per-node gradient alignment scores across all decision points and 8 teachers on BoolQ spans $[-1, +1]$ with a slight positive mean ($+0.027$ for 0.6B, $+0.026$ for 1.7B), demonstrating extreme per-token heterogeneity.

\subsection{Per-path cosine oscillation}

Along individual reasoning paths, the gradient alignment score oscillates token-by-token between positive and negative values, confirming that the teacher's helpfulness is highly local: even on a single path, consecutive tokens can alternate between beneficial and harmful distillation signal.

\subsection{Selective distillation (oracle analysis)}
\label{app:selective}

Given the high per-token heterogeneity, a natural question is: what if we could apply the teacher's gradient \emph{only} at tokens where alignment is positive?
Table~\ref{tab:selective} compares the mean importance-weighted signal under selective strategies that retain only decision points with alignment above a threshold $t$.

\begin{table}[h]
\centering
\caption{Selective distillation (oracle): mean signal (SR range $\times$ cosine), fraction of tokens retained, and fraction of paths that beat full GKD. Filtering to aligned tokens yields $10$--$15\times$ signal improvement using $\sim$50\% of tokens.}
\label{tab:selective}
\small
\begin{tabular}{llccc}
\toprule
\textbf{Setting} & \textbf{Strategy} & \textbf{Mean signal} & \textbf{\% tokens} & \textbf{\% beats GKD} \\
\midrule
\multirow{3}{*}{0.6B BoolQ}
& Full GKD              & $0.007 \pm 0.001$ & 100\%  & --- \\
& Selective ($t{=}0$)   & $0.070 \pm 0.002$ & 51.6\% & 100\% \\
& Selective ($t{=}0.3$) & $0.076 \pm 0.002$ & 46.6\% & 100\% \\
\midrule
\multirow{3}{*}{1.7B BoolQ}
& Full GKD              & $0.006 \pm 0.001$ & 100\%  & --- \\
& Selective ($t{=}0$)   & $0.093 \pm 0.003$ & 51.6\% & 100\% \\
& Selective ($t{=}0.3$) & $0.096 \pm 0.003$ & 49.7\% & 100\% \\
\midrule
\multirow{3}{*}{0.6B MMLU}
& Full GKD              & $0.005 \pm 0.001$ & 100\%  & --- \\
& Selective ($t{=}0$)   & $0.057 \pm 0.002$ & 51.7\% & 100\% \\
& Selective ($t{=}0.3$) & $0.062 \pm 0.002$ & 46.1\% & 100\% \\
\midrule
\multirow{3}{*}{1.7B MMLU}
& Full GKD              & $0.006 \pm 0.001$ & 100\%  & --- \\
& Selective ($t{=}0$)   & $0.077 \pm 0.003$ & 51.2\% & 100\% \\
& Selective ($t{=}0.3$) & $0.081 \pm 0.003$ & 47.9\% & 100\% \\
\bottomrule
\end{tabular}
\end{table}

Simply filtering to tokens with positive alignment ($t = 0$) yields $10$--$15\times$ improvement in mean signal using only $\sim$52\% of tokens, and \emph{every path} benefits.
This is an oracle analysis; at training time, one does not know the true alignment.
However, it establishes a strong upper bound and motivates using the divergence-based predictors identified in Section~\ref{sec:correlations} as approximate filters for alignment-aware training.

\subsection{Teacher ranking by alternative metrics}
\label{app:teacher-alt}

Table~\ref{tab:teacher-ranking-alt} reports teacher rankings using fraction of positively aligned tokens and success-rate-range-weighted cosine.
For 0.6B, both metrics preserve the same hierarchy as mean cosine (Table~\ref{tab:teacher-ranking}).
For 1.7B, the weighted cosine reverses the top ranking: \textsc{Self-Sum-1C} leads over Qwen3-8B, as discussed in Section~\ref{sec:teacher-ranking}.

\begin{table}[h]
\centering
\caption{Teacher ranking by fraction positive and weighted cosine (SR range) on BoolQ.}
\label{tab:teacher-ranking-alt}
\small
\begin{tabular}{lcccc}
\toprule
 & \multicolumn{2}{c}{\textbf{Frac.\ positive}} & \multicolumn{2}{c}{\textbf{Weighted cosine (SR)}} \\
\cmidrule(lr){2-3} \cmidrule(lr){4-5}
\textbf{Teacher} & 0.6B & 1.7B & 0.6B & 1.7B \\
\midrule
\textsc{Self-1C}            & $.526$ & $.515$ & $\mathbf{.120}$ & $.046$ \\
\textsc{Self-1C\,(32B)}       & $\mathbf{.528}$ & $.512$ & $.108$ & $.039$ \\
\textsc{Self-Sum-1C}           & $.522$ & $.527$ & $.114$ & $\mathbf{.088}$ \\
\textsc{Self-Sum-1C1W}         & $.509$ & $.505$ & $.065$ & $.038$ \\
\textsc{Self-1C1W}          & $.511$ & $.501$ & $.043$ & $.017$ \\
\midrule
Qwen3-14B       & $.512$ & $.518$ & $.060$ & $.053$ \\
Qwen3-8B        & $.511$ & $\mathbf{.530}$ & $.062$ & $.072$ \\
Qwen3-4B        & $.512$ & $.520$ & $.053$ & $.060$ \\
\bottomrule
\end{tabular}
\end{table}

The weighted cosine (SR range) amplifies the differences seen in mean cosine: self-distillation teachers for 0.6B achieve weighted cosine $0.108$--$0.120$ vs.\ $0.053$--$0.062$ for external baselines ($2\times$ ratio).
For 1.7B, \textsc{Self-Sum-1C} leads on weighted cosine ($0.088$) followed by Qwen3-8B ($0.072$), showing that at high-stakes decision points the summarized self-distillation context remains competitive even when mean cosine favors the external teacher.
Fraction positive shows smaller but consistent differences in the same direction.

\subsection{Additional teacher ranking plots}
\label{app:teacher-bars}

\begin{figure}[h]
\centering
\includegraphics[width=\textwidth]{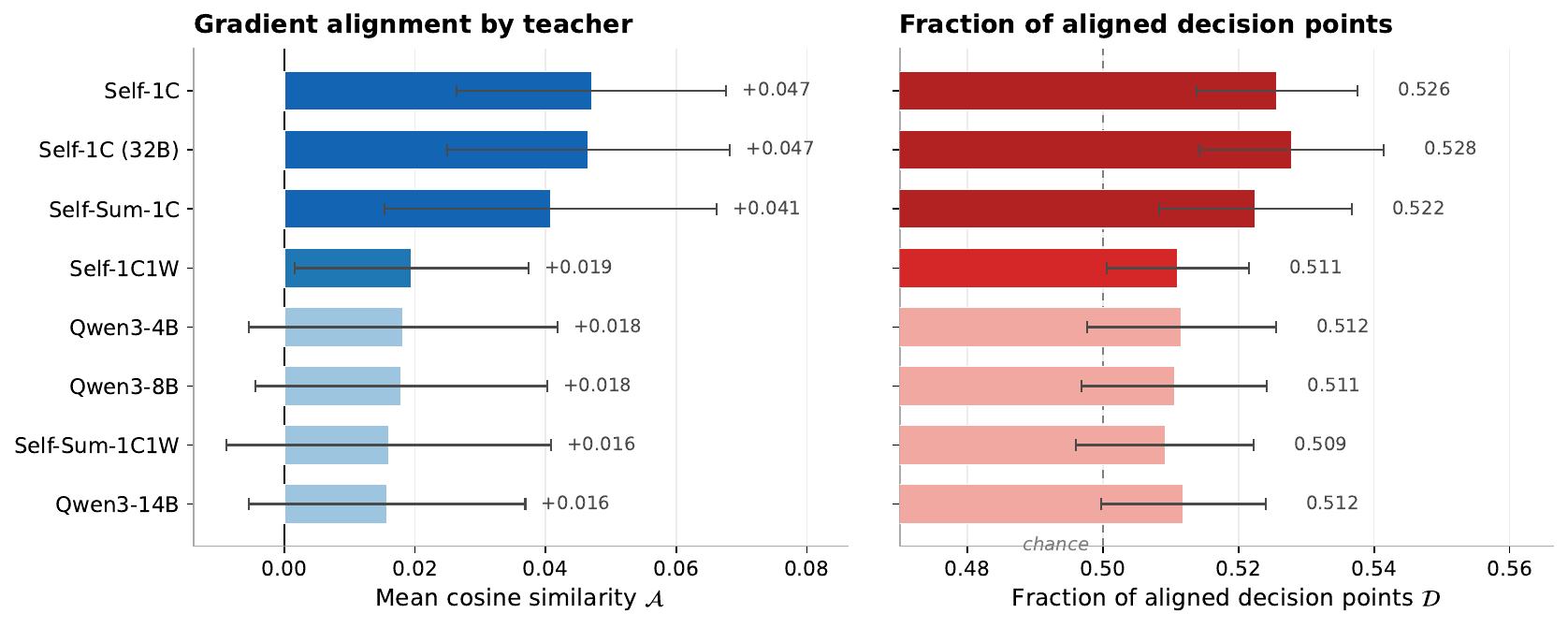}
\captionof{figure}{Qwen3-0.6B, BoolQ}
\vspace{6pt}
\includegraphics[width=\textwidth]{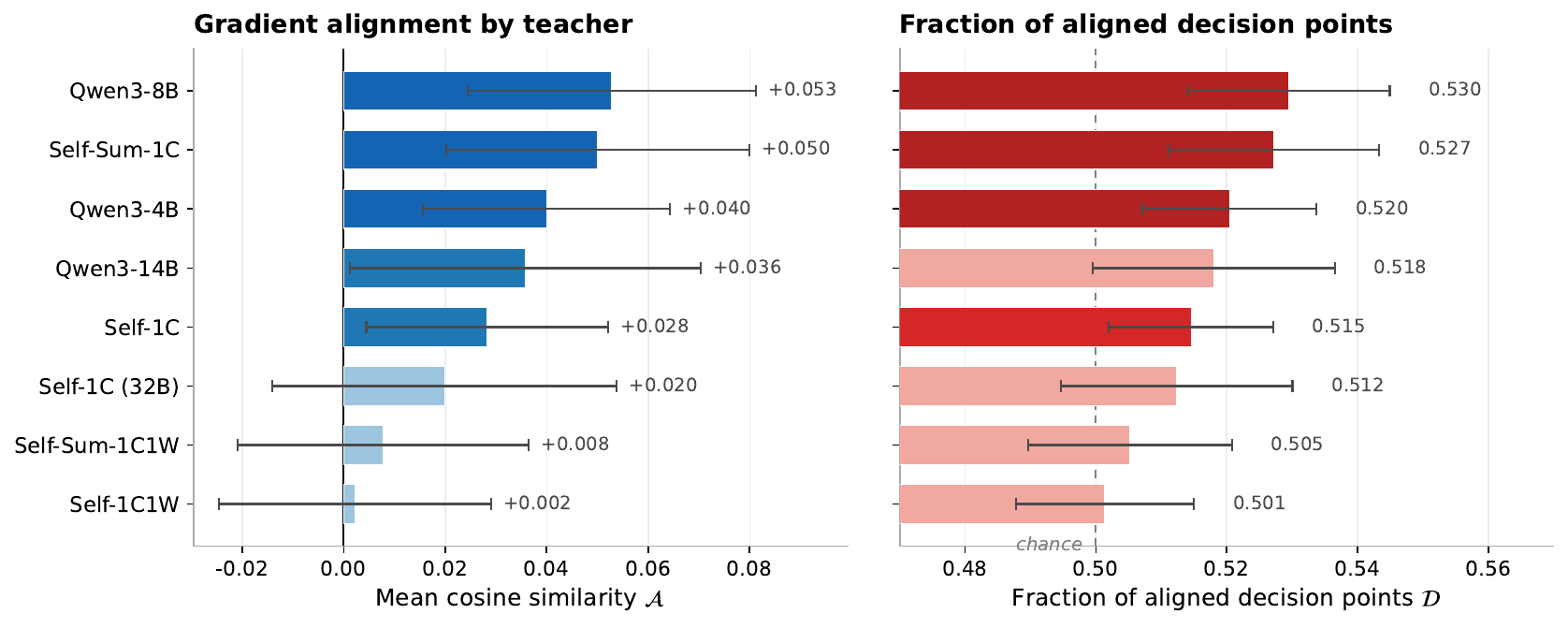}
\captionof{figure}{Qwen3-1.7B, BoolQ}
\vspace{6pt}
\includegraphics[width=\textwidth]{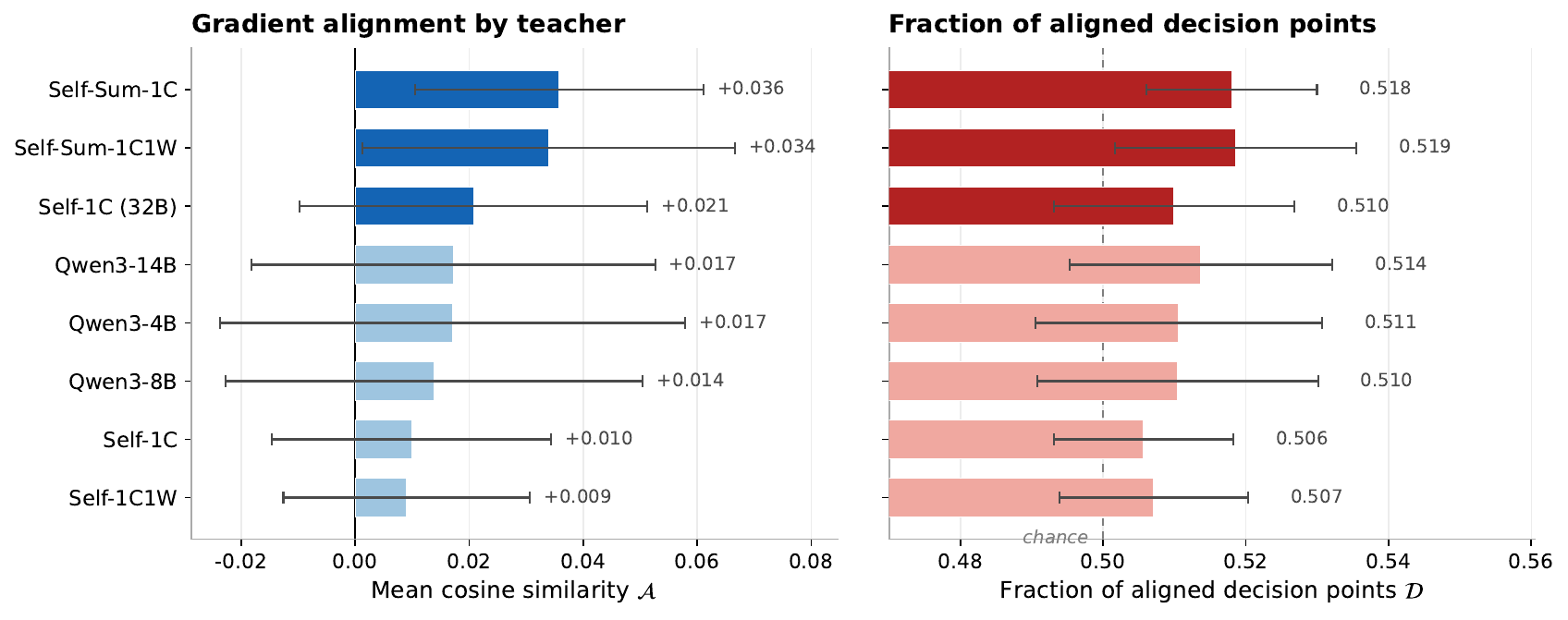}
\captionof{figure}{Qwen3-1.7B, MMLU}
\caption{Teacher ranking by gradient alignment (additional settings). For 1.7B on BoolQ, external teachers (Qwen3-8B) achieve the highest alignment. On MMLU, self-distillation retains an edge for both model scales.}
\label{fig:teacher-bars-appendix}
\end{figure}

\subsection{Full correct vs.\ incorrect breakdown}
\label{app:correct-incorrect-full}

Table~\ref{tab:correct-incorrect-full} extends the main-text results (Figure~\ref{fig:correct-incorrect}) with teacher advantage. The pattern holds across all metrics, with the weighted cosine showing the largest effect sizes.

\begin{table}[h]
\centering
\caption{Full correct vs.\ incorrect path alignment across all metrics.}
\label{tab:correct-incorrect-full}
\small
\begin{tabular}{llcccc}
\toprule
\textbf{Setting} & \textbf{Metric} & \textbf{Correct} & \textbf{Incorrect} & \textbf{$\Delta$} & $p$-value \\
\midrule
\multirow{3}{*}{0.6B BoolQ}
& Mean cosine                & $0.011$ & $0.044$ & $-0.033$ & $7 \times 10^{-8}$ \\
& Weighted cosine (SR range) & $0.046$ & $0.110$ & $-0.065$ & $8 \times 10^{-10}$ \\
& Teacher advantage          & $0.002$ & $0.008$ & $-0.006$ & $< 10^{-19}$ \\
\midrule
\multirow{3}{*}{1.7B BoolQ}
& Mean cosine                & $0.001$ & $0.058$ & $-0.056$ & $2 \times 10^{-9}$ \\
& Weighted cosine (SR range) & $0.010$ & $0.093$ & $-0.083$ & $7 \times 10^{-9}$ \\
& Teacher advantage          & $-0.002$ & $0.011$ & $-0.013$ & $< 10^{-33}$ \\
\midrule
\multirow{3}{*}{0.6B MMLU}
& Mean cosine                & $0.009$ & $0.048$ & $-0.039$ & $0.0001$ \\
& Weighted cosine (SR range) & $0.021$ & $0.118$ & $-0.097$ & $< 10^{-11}$ \\
& Teacher advantage          & $0.002$ & $0.005$ & $-0.004$ & $< 10^{-14}$ \\
\midrule
\multirow{3}{*}{1.7B MMLU}
& Mean cosine                & $0.012$ & $0.028$ & $-0.016$ & $0.123$ \\
& Weighted cosine (SR range) & $0.034$ & $0.098$ & $-0.063$ & $< 10^{-5}$ \\
& Teacher advantage          & $0.001$ & $0.010$ & $-0.009$ & $< 10^{-14}$ \\
\bottomrule
\end{tabular}
\end{table}

\subsection{Within-path correlation details}
\label{app:correlations}

\begin{table}[h]
\centering
\caption{Within-path Spearman correlations between single-rollout features and the gradient alignment score. Top: averaged over all 8 teachers. Bottom: restricted to a teacher per setting.}
\label{tab:correlations}
\small
\begin{tabular}{lcccc}
\toprule
 & \multicolumn{2}{c}{\textbf{Qwen3-0.6B}} & \multicolumn{2}{c}{\textbf{Qwen3-1.7B}} \\
\cmidrule(lr){2-3} \cmidrule(lr){4-5}
\textbf{Feature} & BoolQ & MMLU & BoolQ & MMLU \\
\midrule
\multicolumn{5}{l}{\emph{All teachers combined}} \\
Depth (normalized)    & $+.042$ & $+.035$ & $-.007$ & $+.010$ \\
KL($\pi_\theta \| \pi_{\mathrm{te}}$) & $+.029$ & $+.028$ & $+.009$ & $+.030$ \\
JS divergence         & $+.030$ & $+.028$ & $+.006$ & $+.022$ \\
$L_2$ distance        & $+.031$ & $+.034$ & $+.006$ & $+.023$ \\
Cosine($\pi_\theta, \pi_{\mathrm{te}}$) & $-.029$ & $-.026$ & $-.003$ & $-.020$ \\
\midrule
\multicolumn{5}{l}{\emph{With a teacher only}} \\
 & \footnotesize\textsc{Self-1C} & \footnotesize\textsc{Self-1C\,(32B)} & \footnotesize\textsc{Self-Sum-1C} & \footnotesize\textsc{Self-Sum-1C} \\
Depth (normalized)    & $+.042$ & $+.026$ & $+.000$ & $+.028$ \\
KL($\pi_\theta \| \pi_{\mathrm{te}}$) & $+.044$ & $+.000$ & $+.011$ & $+.028$ \\
JS divergence         & $+.041$ & $-.002$ & $+.009$ & $+.020$ \\
$L_2$ distance        & $+.042$ & $+.013$ & $+.009$ & $+.020$ \\
Cosine($\pi_\theta, \pi_{\mathrm{te}}$) & $-.034$ & $-.005$ & $-.005$ & $-.019$ \\
\bottomrule
\end{tabular}
\end{table}

The divergence$\to$alignment pattern is consistent for 0.6B and strengthens when restricting to the best teacher, but weakens for 1.7B BoolQ regardless of teacher choice.
The positive sign means alignment is higher where the teacher disagrees with the student: low-divergence tokens offer little useful signal.
However, high divergence is necessary but not sufficient: many high-divergence tokens still have negative alignment, since the teacher's confidence can point toward failure as easily as success.
Depth correlates positively for 0.6B ($\rho \approx +0.04$), reflecting that early tokens are templatic while later tokens involve reasoning where the teacher's contextual advantage emerges; this effect vanishes for 1.7B.
On AIME (longer traces), the divergence pattern persists but depth becomes weakly negative ($\rho \approx -0.03$ on hard questions), suggesting that on complex math the teacher's advantage does not grow with depth.

\subsection{Per-teacher correct vs.\ incorrect breakdown}
\label{app:per-teacher-delta}

Table~\ref{tab:per-teacher-delta} breaks down the correct-vs-incorrect gap by teacher on BoolQ.
For 0.6B, \textsc{Self-1C} is unique: it achieves nearly equal alignment on both correct and incorrect paths ($\Delta \approx 0$), providing a uniformly helpful gradient regardless of path outcome.
All other teachers show the typical pattern of higher alignment on incorrect paths; for 1.7B this holds without exception.

\begin{table}[h]
\centering
\caption{Per-teacher correct vs.\ incorrect mean cosine on BoolQ.}
\label{tab:per-teacher-delta}
\small
\begin{tabular}{lcccccc}
\toprule
 & \multicolumn{3}{c}{\textbf{Qwen3-0.6B}} & \multicolumn{3}{c}{\textbf{Qwen3-1.7B}} \\
\cmidrule(lr){2-4} \cmidrule(lr){5-7}
\textbf{Teacher} & \textbf{Corr.} & \textbf{Incorr.} & $\Delta$ & \textbf{Corr.} & \textbf{Incorr.} & $\Delta$ \\
\midrule
\textsc{Self-1C}       & $0.047$  & $0.047$  & $\approx 0$ & $-0.006$ & $0.062$  & $-0.068$ \\
\textsc{Self-1C\,(32B)}  & $0.030$  & $0.063$  & $-0.034$ & $-0.027$ & $0.067$  & $-0.094$ \\
\textsc{Self-Sum-1C}      & $0.023$  & $0.059$  & $-0.036$ & $0.021$  & $0.079$  & $-0.057$ \\
\textsc{Self-1C1W}     & $0.010$  & $0.029$  & $-0.018$ & $-0.013$ & $0.017$  & $-0.030$ \\
\textsc{Self-Sum-1C1W}    & $-0.005$ & $0.037$  & $-0.042$ & $-0.026$ & $0.042$  & $-0.068$ \\
\midrule
Qwen3-14B  & $-0.006$ & $0.037$  & $-0.043$ & $0.017$  & $0.055$  & $-0.039$ \\
Qwen3-8B   & $-0.012$ & $0.048$  & $-0.059$ & $0.026$  & $0.080$  & $-0.055$ \\
Qwen3-4B   & $0.001$  & $0.035$  & $-0.035$ & $0.020$  & $0.060$  & $-0.040$ \\
\bottomrule
\end{tabular}
\end{table}

\subsection{Teacher Context Generation and Screening Details}
\label{app:context_screening}

To investigate the impact of in-context demonstrations on performance, we design a screening pipeline that measures the impact of various forms of context on pass rates across difficulty levels. For sourced context in summarized demonstrations, we use Qwen3-32B to process demonstrations. For the rest, we use the same model as the student. All models use thinking mode enabled. We screen questions from two benchmarks: MMLU~\citep{hendrycks2021mmlu} (500 questions, multiple-choice) and BoolQ~\citep{clark2019boolq} (500 questions, yes/no). For each question we sample $G{=}32$ rollouts at temperature $\tau{=}1.0$ and compute the pass rate as the fraction of rollouts producing a correct answer.

\paragraph{Context Generation.}
Before screening, for each question, we generate demonstration responses by sampling from the context source model at temperature $\tau_{\text{demo}}{=}0.7$. Each response is verified against the ground-truth answer and classified as correct or incorrect. Generation retries up to 60 trials per question to collect the required number of demonstrations.

We evaluate seven context configurations, organized into three families:
\begin{enumerate}
    \item \textbf{Raw demonstrations.} Correct and/or incorrect responses are prepended verbatim to the prompt. Variants: 1~correct (\textsc{Self-1C}), 1~correct + 1~wrong (\textsc{Self-1C1W}), and 3~correct (\textsc{Self-3C}). Additionally, \textsc{Self-1C\,(32B)} uses a single correct demonstration generated by Qwen3-32B rather than the student.
    \item \textbf{Summarized demonstrations.} Qwen3-32B condenses each response to its key reasoning steps and final answer. Variants: 1~summarized correct (\textsc{Self-Sum-1C}) and 1~summarized correct + 1~wrong (\textsc{Self-Sum-1C1W}).
\end{enumerate}

\paragraph{Context Injection.}
All context is injected by prepending it to the user message before applying the chat template. Figures~\ref{fig:prompt_baseline}--\ref{fig:prompt_demo1c1w} show the exact prompt structure for each context mode.

\input{prompts.tex}
\paragraph{Difficulty Binning and Filtering.}
We assign each question to a difficulty bin based on its \emph{baseline} (no-context) pass rate: \emph{easy} ($p \geq 0.8$), \emph{medium} ($0.25 \leq p < 0.8$), \emph{hard} ($0 < p < 0.25$), and \emph{extremely hard} ($p = 0$). Extremely hard questions are excluded from the primary analysis as a zero baseline provides no signal for measuring improvement during offline gradient analysis.

To ensure fair comparison across context variants, we restrict the analysis to questions for which the context generation process successfully produced at least one correct demonstration under \emph{both} the self-context and 32B-context conditions. Table~\ref{tab:context_filter} reports the number of questions retained after this filter. It is worth to emphasize that the \textbf{questions where no correct response could be generated for either source are excluded}, ensuring that observed differences reflect the \emph{quality} of injected context rather than its \emph{availability}.

\input{tables/screening_details}

Several patterns emerge across all models and benchmarks. First, even a single correct demonstration (\textsc{Self-1C}) as teacher context produces dramatic improvements for teacher accuracy. Second, including wrong demonstrations alongside correct ones (\textsc{Self-1C1W}) consistently hurts performance relative to correct-only variants, and sometimes even degrades below the no-context baseline. Third, the gap between self-generated and 32B-generated demonstrations (\textsc{Self-1C} vs.\ \textsc{Self-1C\,(32B)}) is small. Lastly, we also provide difficulty breakdown for accuracy changes in screening questions based on context variations in Tables~\ref{tab:breakdown}. We observe that the accuracy improvements for medium and easy questions (for baseline) are more significant compared to the improvements for hard questions, which suggests that especially the smaller models may still tend to generate wrong responses even when correct demonstrations are provided in its context.

\input{tables/screening_results}
\input{tables/screening_breakdown}

\applefootnote{ \textcolor{textgray}{\sffamily Apple and the Apple logo are trademarks of Apple Inc., registered in the U.S. and other countries and regions.}}

\end{document}

%% file: apple_preamble.tex
\usepackage{amsmath}
\usepackage{enumerate}
\usepackage{algorithm}
\usepackage{algpseudocode}
\usepackage{amsfonts}
\usepackage{amsthm}
\usepackage{cleveref}
\usepackage{diagbox}
\usepackage{colortbl}
\usepackage{amssymb}
\usepackage{xspace}
\usepackage{wrapfig}
\usepackage{adjustbox}
\usepackage{tabularx}
\usepackage{booktabs}
\usepackage{mathtools}
\usepackage{tikz}
\usepackage{enumitem}
\usepackage{silence}
\usepackage{dsfont}
\usepackage[table]{xcolor}
\usepackage[dvipsnames]{xcolor}
\usepackage{multirow}
\usepackage{makecell}
\usepackage{xfakebold}
\input{math_commands}

\usepackage{nicefrac}
\usepackage[textsize=tiny]{todonotes}

\definecolor{textgray}{HTML}{6E6E73}
\usetikzlibrary{positioning, calc}
\usetikzlibrary{decorations.pathmorphing}
\usetikzlibrary{arrows.meta, decorations.pathreplacing, backgrounds}

\definecolor{correctgreen}{RGB}{0,145,65}
\definecolor{wrongred}{RGB}{175,30,30}
\definecolor{alignblue}{RGB}{55,120,195}
\definecolor{misalignred}{RGB}{195,55,55}
\definecolor{neutralgray}{RGB}{140,140,140}
\definecolor{highlightyellow}{RGB}{255,252,210}

\definecolor{systembg}{HTML}{F0F4FF}
\definecolor{systemframe}{HTML}{6B7FD7}
\definecolor{userbg}{HTML}{F6FFF6}
\definecolor{userframe}{HTML}{4CAF50}
\definecolor{demobg}{HTML}{FFF8F0}
\definecolor{demoframe}{HTML}{E8913A}
\definecolor{promptgray}{HTML}{F7F7F8}
\definecolor{promptborder}{HTML}{D1D5DB}
\definecolor{accentblue}{HTML}{3B82F6}
\definecolor{promptgreen}{HTML}{2E7D32}
\definecolor{promptred}{HTML}{C62828}
\definecolor{summcolor}{HTML}{E65100}
\definecolor{lessoncolor}{HTML}{00796B}

\makeatletter
\patchcmd{\wrong@fontshape}{\@gobbletwo}{}{}{}
\makeatother
\numberwithin{equation}{section}
\makeatletter
\AtBeginDocument{
  \urlstyle{sf}
  
}
\makeatother

\definecolor{light}{RGB}{125, 125, 125}
\crefname{tcb@cnt@pbox}{code}{code}
\Crefname{tcb@cnt@pbox}{Code}{Code}
\crefname{assumption}{assumption}{assumption}
\Crefname{assumption}{Assumption}{Assumptions}

\newtcolorbox[auto counter]{pbox}[2][]{
  colback=white,
  title=Code~\thetcbcounter: #2,
  #1,fonttitle=\sffamily,
  fontupper=\sffamily,
  arc=2pt,
  colframe=bgcolor,
  coltitle=fgcolor,
  colbacktitle=bgcolor,
  toptitle=0.25cm,
  bottomtitle=0.125cm
}

\makeatletter
\newcommand\applefootnote[1]{%
  \begingroup
  \renewcommand\thefootnote{}%
  \renewcommand\@makefntext[1]{\noindent##1}%
  \footnote{#1}%
  \addtocounter{footnote}{-1}%
  \endgroup
}
\makeatother

\definecolor{cverbbg}{gray}{0.90}

%% file: math_commands.tex
\usepackage{amsmath,amsfonts,bm}

\def\eqref#1{equation~\ref{#1}}

\def\1{\bm{1}}

\DeclareMathAlphabet{\mathsfit}{\encodingdefault}{\sfdefault}{m}{sl}
\SetMathAlphabet{\mathsfit}{bold}{\encodingdefault}{\sfdefault}{bx}{n}



%% file: figures/figure1.tex

\definecolor{fig@nodebg}{RGB}{245,245,245}
\definecolor{fig@prefixbg}{RGB}{230,235,245}
\definecolor{fig@studentblue}{RGB}{70,120,200}
\definecolor{fig@teacherorange}{RGB}{200,120,30}

\tikzset{
  fig1/tok/.style={
    rectangle, rounded corners=2pt,
    minimum height=0.55cm,
    font=\small\ttfamily, inner sep=4pt,
    draw=gray!50, fill=fig@nodebg
  },
  fig1/tok_prefix/.style={fig1/tok, fill=fig@prefixbg, draw=gray!40},
  fig1/tok_student/.style={fig1/tok, fill=fig@studentblue!16, draw=fig@studentblue!55},
  fig1/tok_student_bold/.style={fig1/tok_student, font=\small\ttfamily\bfseries},
  fig1/tok_teacher/.style={fig1/tok, fill=fig@teacherorange!18, draw=fig@teacherorange!60},
  fig1/tok_teacher_bold/.style={fig1/tok_teacher, font=\small\ttfamily\bfseries},
  fig1/tree_edge/.style={-{Stealth[length=4pt]}, gray!50, line width=0.7pt},
  fig1/blabel/.style={
    font=\scriptsize\itshape, text=gray!65, align=center, inner sep=1pt
  },
}

\begin{tikzpicture}[>=Stealth]

\node[
  draw=black!40, rounded corners=5pt, fill=highlightyellow,
  text width=16.5cm, align=center, inner sep=7pt, font=\normalsize
] (problem) at (7.0, 0) {
  \textbf{Problem:} A bookshelf has 3 shelves. Each shelf holds 4 books. If 5 books are removed, how many remain? \quad \textbf{Answer: 7}
};

\node[fig1/tok_prefix, below=1.8cm of problem.south west,
      anchor=north west, xshift=0.2cm] (p1) {3 shelves};

\node[minimum size=0pt, inner sep=0pt]
    (b0) at ($(p1.east)+(0.45,0)$) {};
\fill[gray!55] (b0) circle (1.5pt);

\node[fig1/tok_prefix, right=18pt of b0] (p2a) {4};
\node[fig1/tok_prefix, right=4pt of p2a] (p2b) {books};
\node[fig1/tok_prefix, right=4pt of p2b] (p2c) {each};
\node[fig1/tok_prefix, right=4pt of p2c] (pdots) {$\cdots$};

\draw[gray!50, line width=0.6pt] (p1.east) -- (b0);
\draw[fig1/tree_edge] (b0) -- (p2a.west);
\foreach \a/\b in {p2a/p2b, p2b/p2c, p2c/pdots}
  \draw[fig1/tree_edge] (\a) -- (\b);

\node[font=\small\itshape, text=gray!55,
      left=4pt of p1.west, anchor=east] {student answer:};


\node[minimum size=0pt, inner sep=0pt]
    (b1) at ($(pdots.east)+(0.45,0)$) {};
\fill[gray!55] (b1) circle (1.5pt);

\node[fig1/tok_student, right=18pt of b1] (m1) {so,};
\node[fig1/tok_student, right=4pt of m1]  (m2) {total is};
\node[fig1/tok_student, right=4pt of m2]  (m3) {3};

\draw[gray!50, line width=0.6pt] (pdots.east) -- (b1);
\draw[fig1/tree_edge] (b1) -- (m1.west);
\foreach \a/\b in {m1/m2,m2/m3}
  \draw[fig1/tree_edge] (\a) -- (\b);

\node[minimum size=0pt, inner sep=0pt]
    (b2) at ($(m3.east)+(0.45,0)$) {};
\fill[gray!55] (b2) circle (1.5pt);

\node[fig1/tok_student_bold, right=18pt of b2] (n1) {$+4{=}7.$};

\draw[gray!50, line width=0.6pt] (m3.east) -- (b2);
\draw[fig1/tree_edge] (b2) -- (n1.west);

\node[minimum size=0pt, inner sep=0pt]
    (b3) at ($(n1.east)+(1.10,0)$) {};
\fill[gray!55] (b3) circle (1.5pt);

\node[fig1/tok_student, right=18pt of b3] (n2) {Rem.\ 5:};
\node[fig1/tok_student_bold, right=4pt of n2] (n3)
    {${=}2\,$\textcolor{wrongred}{$\boldsymbol{\times}$}};

\draw[gray!50, line width=0.6pt] (n1.east) -- (b3);
\draw[fig1/tree_edge] (b3) -- (n2.west);
\draw[fig1/tree_edge] (n2) -- (n3);


\node[fig1/tok_teacher, anchor=west]
    (s0) at ($(b0)+(1.0, 1.45)$) {four};
\node[fig1/tok_student, right=4pt of s0] (s0b) {$\cdots$};

\draw[fig1/tree_edge] (b0) to[out=78, in=200] (s0.west);
\draw[fig1/tree_edge] (s0) -- (s0b);

\node[fig1/tok_teacher, anchor=west]
    (s1) at ($(b1)+(1.0, 1.45)$) {therefore};
\node[fig1/tok_student, right=4pt of s1] (s1b) {$\cdots$};

\draw[fig1/tree_edge] (b1) to[out=78, in=200] (s1.west);
\draw[fig1/tree_edge] (s1) -- (s1b);

\node[fig1/tok_teacher_bold, anchor=west]
    (s2) at ($(b2)+(1.0, 1.45)$) {${\times}4{=}12.$};
\node[fig1/tok_student_bold, right=4pt of s2] (s2c)
    {${=}7\,$\textcolor{correctgreen}{$\boldsymbol{\checkmark}$}};

\draw[fig1/tree_edge] (b2) to[out=78, in=200] (s2.west);
\draw[fig1/tree_edge] (s2) -- (s2c);

\node[fig1/tok_teacher, anchor=west]
    (s3) at ($(b3)+(1.5, 1.45)$) {$12{-}5{:}$};
\node[fig1/tok_student_bold, right=4pt of s3] (s3b)
    {${=}7\,$\textcolor{correctgreen}{$\boldsymbol{\checkmark}$}};

\draw[fig1/tree_edge] (b3) to[out=78, in=200] (s3.west);
\draw[fig1/tree_edge] (s3) -- (s3b);

\node[fig1/blabel] at ($(b0)+(1.30, 0.72)$) {notation\\(stylistic)};
\node[fig1/blabel] at ($(b1)+(1.30, 0.72)$) {word choice\\(stylistic)};
\node[fig1/blabel] at ($(b2)+(1.30, 0.72)$) {reasoning\\(critical)};
\node[fig1/blabel] at ($(b3)+(1.60, 0.72)$) {error recovery\\(post-hoc)};

\node[fig1/tok_student, minimum width=0.5cm, minimum height=0.32cm, inner sep=0pt,
      anchor=north west] (lbox_s) at ($(p1.north west)+(-1.5, 1.40)$) {};
\node[font=\small, anchor=west, right=4pt of lbox_s] (ltxt_s) {Student token};

\node[fig1/tok_teacher, minimum width=0.5cm, minimum height=0.32cm, inner sep=0pt,
      below=5pt of lbox_s] (lbox_t) {};
\node[font=\small, anchor=west, right=4pt of lbox_t] {Teacher suggestion};

\end{tikzpicture}

%% file: method.tex
\section{Methodology}
\label{sec:method}

\subsection{Not all teacher guidance is useful}
\label{sec:method-problem}

At each token position, the teacher's distribution may differ from the student's for many reasons: it may prefer a stylistic variant, it may encourage the student along a productive reasoning path, or it may redirect the computation entirely toward a different continuation.
The core difficulty is that \emph{one cannot distinguish these cases from the teacher's probability alone}.
A token where the teacher and student assign substantially different probabilities could reflect any of these, and only some improve the student's chance of reaching a correct answer (cf.\ Figure~\ref{fig:bookshelf}).
To tell them apart, we need to connect the teacher's token-level signal to downstream outcomes.

We do this by decomposing the generation process into a \emph{generation tree}: given $G$ trajectories sampled from the student $\pi_\theta$ on a prompt $q$, each trajectory shares prefixes with others, forming a tree where each node $u$ corresponds to a token position and each edge corresponds to a next-token choice.
At each node $u$, we observe which next tokens were chosen across rollouts and which of those rollouts ultimately reached a correct answer.
This gives us an empirical estimate of the \emph{success probability} $\hat{P}_{\mathrm{succ}}^k$: the probability of reaching a correct answer after choosing token $k$ at node $u$.

With this quantity in hand, we can ask precisely: does the teacher push probability mass toward high-$P_{\mathrm{succ}}$ tokens, or away from them?

Independent of any training algorithm, a teacher is good at node $u$ if it places more mass on success-leading tokens than the student does.
We define the \textbf{teacher advantage}:
\begin{equation}
  \label{eq:advantage}
  \mathrm{Adv}(u) = \sum_{k \in \mathcal{S}_u} P_{\mathrm{te}}^k \, \hat{P}_{\mathrm{succ}}^k
  - \sum_{k \in \mathcal{S}_u} P_\theta^k \, \hat{P}_{\mathrm{succ}}^k,
\end{equation}
where $\mathcal{S}_u$ is the set of tokens with sufficient visit counts at node $u$, $P_{\mathrm{te}}^k$ is the teacher's probability of token $k$, $P_\theta^k$ is the student's probability of token $k$, and probabilities are renormalized over $\mathcal{S}_u$.
A positive advantage means the teacher ``knows better'' at this node; a negative advantage means following it would hurt.

But a good teacher is not sufficient: you also need an algorithm that translates the teacher's knowledge into a useful gradient, and different algorithms (GKD~\citep{agarwal2024gkd}, single-sample estimators~\citep{thinking2025distillation}, MiniLLM~\citep{gu2024minillm}) use the teacher differently, producing very different gradients from the same teacher.
To evaluate any (teacher, algorithm) pair, we need an \textbf{ideal reference}: the gradient that would optimally improve the student's success probability at each node.

\subsection{The ideal reference gradient}
\label{sec:method-ideal}

At each node $u$, the ideal local objective is to maximize the student's probability of reaching a correct answer from this point:
\begin{equation}
  \label{eq:ideal}
  L_{\mathrm{ideal}}(u) = \sum_{k} P_\theta^k \, P_{\mathrm{succ}}^k.
\end{equation}
This is the expected success rate under the student's current distribution at node $u$.
The gradient of this objective with respect to the student's logit $z_j$ at this node is obtained via the softmax Jacobian $\partial P_\theta^k / \partial z_j = P_\theta^k(\delta_{kj} - P_\theta^j)$, where $\delta_{kj}$ is the Kronecker delta ($1$ if $k=j$, $0$ otherwise):
\begin{align}
  \frac{\partial L_{\mathrm{ideal}}}{\partial z_j}
  &= \sum_k P_{\mathrm{succ}}^k \cdot P_\theta^k(\delta_{kj} - P_\theta^j) \nonumber \\
  &= P_\theta^j \, P_{\mathrm{succ}}^j - P_\theta^j \sum_k P_\theta^k P_{\mathrm{succ}}^k \nonumber \\
  &= P_\theta^j \left( P_{\mathrm{succ}}^j - \bar{P}_{\mathrm{succ}} \right),
  \label{eq:ideal-grad}
\end{align}
where $\bar{P}_{\mathrm{succ}} = \sum_k P_\theta^k P_{\mathrm{succ}}^k$ is the student's current expected success at this node.
This gradient increases logit $z_j$ when token $j$ leads to success more often than average, and decreases it otherwise.
This is our reference: the direction in which the student's logits should move to maximally improve its chance of success at this node.

\paragraph{Dr.\ GRPO recovers this gradient in expectation.}
A natural question is whether any existing training objective already computes this ideal gradient.
Dr.\ GRPO~\citep{liu2025understanding} is a variant of GRPO that removes the per-trajectory length normalization $1/|o_i|$.
The full GRPO objective includes an importance ratio $\pi_\theta/\pi_{\theta_\mathrm{old}}$, a KL penalty, and division by trajectory length.
After marginalizing the importance ratio, dropping the KL penalty (small $\beta$), and removing length normalization, the expected objective at node $u$ reduces to $\sum_k P_\theta^k \cdot P_{\mathrm{succ}}^k$ up to constants independent of $\theta$ (see Appendix~\ref{app:derivations} for the full derivation).
The empirical per-sample gradient at node $u$ is:
\begin{equation}
  \frac{\partial \hat{L}_{\mathrm{DrGRPO}}}{\partial z_j}\bigg|_u
  = \frac{1}{N_u}\sum_{i:\, o_i \ni u} A_i \left(\delta_{r_i, j} - P_\theta^j\right),
\end{equation}
where $A_i = (R_i - \bar{R}) / \mathrm{std}(R)$ is the normalized advantage, $r_i$ is the token chosen by rollout $i$, and $N_u$ is the number of rollouts through $u$.
In expectation, this is proportional to the ideal gradient (Equation~\ref{eq:ideal-grad}):
\begin{equation}
  \mathbb{E}\left[\nabla_{z_j} L_{\mathrm{DrGRPO}}\big|_u\right]
  \;\propto\; P_\theta^j (P_{\mathrm{succ}}^j - \bar{P}_{\mathrm{succ}}).
\end{equation}
This connection motivates using Equation~\eqref{eq:ideal-grad} as our oracle reference: it is what reward-based training would converge toward at each node, given sufficient rollouts.
Standard GRPO's $1/|o_i|$ factor couples the advantage to trajectory length, preventing this clean per-node decomposition.

In practice, we compute the ideal gradient \emph{directly} from empirical $\hat{P}_{\mathrm{succ}}^k$ estimates at each node, not from per-sample advantage terms.
This is more accurate than the finite-sample Dr.\ GRPO estimator and avoids the noise of individual trajectory rewards.

\subsection{Distillation gradients}
\label{sec:method-distill}

We now derive the gradient that each distillation algorithm produces at node $u$.

\paragraph{GKD (Generalized Knowledge Distillation).}
GKD~\citep{agarwal2024gkd} minimizes the forward KL from student to teacher at each node:
\begin{equation}
  L_{\mathrm{GKD}}(u) = \mathrm{KL}(\pi_\theta \| \pi_{\mathrm{te}}) = \sum_k P_\theta^k \left(\log P_\theta^k - \log P_{\mathrm{te}}^k\right)
\end{equation}
Defining the per-token log-ratio $\ell_k = \log P_\theta^k - \log P_{\mathrm{te}}^k$, the gradient is:
\begin{align}
  \frac{\partial L_{\mathrm{GKD}}}{\partial z_j}
  &= \sum_k \frac{\partial}{\partial z_j}\left[ P_\theta^k \, \ell_k \right] \nonumber \\
  &= \sum_k \ell_k \cdot P_\theta^k(\delta_{kj} - P_\theta^j) + \sum_k P_\theta^k \cdot \frac{\partial \ell_k}{\partial z_j}
\end{align}
The second sum contributes $\sum_k P_\theta^k \cdot (\delta_{kj} - P_\theta^j) = P_\theta^j - P_\theta^j = 0$ (the softmax Jacobian sums to zero), so:
\begin{equation}
  \frac{\partial L_{\mathrm{GKD}}}{\partial z_j} = P_\theta^j(\ell_j - \bar{\ell})
  \label{eq:gkd-grad}
\end{equation}
where $\bar{\ell} = \sum_k P_\theta^k \ell_k = \mathrm{KL}(\pi_\theta \| \pi_{\mathrm{te}})$.
Since we minimize this KL, the distillation gradient (with sign flip) is $-P_\theta^j(\ell_j - \bar\ell)$, which pushes logits toward tokens where the teacher assigns relatively higher probability.

\paragraph{Single-sample GKD estimator.}
\citet{thinking2025distillation} propose an importance-weighted estimator requiring only the sampled token.
For rollout $i$ choosing token $r_i$ at node $u$, the per-sample gradient is:
\begin{equation}
  \nabla_{z_j} \ell_i = \left(\log P_{\mathrm{te}}^{r_i} - \log P_\theta^{r_i} - 1\right) \cdot (\delta_{r_i, j} - P_\theta^j)
\end{equation}
In expectation this recovers $-P_\theta^j(\ell_j - \bar{\ell})$, the GKD gradient with opposite sign (the $-1$ baseline vanishes; see Appendix~\ref{app:derivations}).

\paragraph{MiniLLM.}
MiniLLM~\citep{gu2024minillm} uses a REINFORCE-style gradient with trajectory-level reward-to-go:
\begin{equation}
  \nabla_{z_j} \ell_i = -(\delta_{r_i, j} - P_\theta^j) \cdot \left(\sum_{t' \geq t} \left(\log P_{\mathrm{te}}^{o_{t'}} - \log P_\theta^{o_{t'}}\right) - 1\right)
\end{equation}
This couples the gradient at node $u$ to all downstream nodes. The local gradient still takes the form $P_\theta^j(f_j - \bar{f})$ in expectation, but $f_k$ is now trajectory-dependent rather than purely local (see Appendix~\ref{app:derivations}).

\paragraph{Summary.}
All four methods produce per-node gradients of the form:
\begin{equation}
  \frac{\partial L}{\partial z_j} = P_\theta^j(f_j - \bar{f}), \qquad \bar{f} = \sum_k P_\theta^k f_k
  \label{eq:unified}
\end{equation}
with $f_k = P_{\mathrm{succ}}^k$ for Dr.\ GRPO, $f_k = \pm(\log P_\theta^k - \log P_{\mathrm{te}}^k)$ for GKD (and its single-sample estimator), and a trajectory-dependent reward-to-go for MiniLLM.
Because they share this structure, we can compare their \emph{directions} via cosine similarity.
A consequence of the shared $P_\theta^j$ factor is that the gradient magnitude for any token is gated by the student's current probability: even if the teacher identifies a high-success token, the update is small when $P_\theta^j$ is small.
The teacher can amplify tokens the student already partially believes in, but has limited ability to inject entirely new continuations in a single step.

\subsection{The gradient alignment score}
\label{sec:method-score}

We define the \textbf{gradient alignment score} at node $u$ as the cosine similarity between the ideal gradient (Equation~\ref{eq:ideal-grad}) and the distillation gradient (e.g., GKD):
\begin{equation}
  \mathrm{Align}(u) = \cos\!\left(\mathbf{g}^{\mathrm{ideal}}_u,\, \mathbf{g}^{\mathrm{D}}_u\right)
  = \frac{\sum_{j \in \mathcal{S}_u} g_j^{\mathrm{ideal}} \, g_j^{\mathrm{D}}}
         {\|\mathbf{g}^{\mathrm{ideal}}\|_{\mathcal{S}_u} \; \|\mathbf{g}^{\mathrm{D}}\|_{\mathcal{S}_u}}
  \label{eq:alignment}
\end{equation}
where $\mathbf{g}^{\mathrm{ideal}}_u$ is the ideal gradient computed directly from empirical $\hat{P}_{\mathrm{succ}}^k$ and $\mathbf{g}^{\mathrm{D}}_u$ is the distillation gradient vector, both restricted to $\mathcal{S}_u$ (the set of tokens with sufficient visit counts at node $u$).
The restriction is necessary because $P_{\mathrm{succ}}^k$ is only reliably estimated for tokens that have been sampled enough times.

The score ranges from $-1$ to $+1$:
\begin{itemize}[leftmargin=*, itemsep=2pt]
  \item $\mathrm{Align}(u) > 0$: the distillation gradient pushes toward tokens that lead to success. The teacher + algorithm combination is \emph{helpful} at this node.
  \item $\mathrm{Align}(u) \approx 0$: the distillation gradient is orthogonal to the reward signal. The teacher's guidance is neither helpful nor harmful; it wastes gradient budget on irrelevant directions (e.g., stylistic preferences).
  \item $\mathrm{Align}(u) < 0$: the distillation gradient pushes toward tokens that lead to failure. The teacher + algorithm combination is \emph{actively harmful} at this node.
\end{itemize}

This score answers the question posed in Section~\ref{sec:method-problem}: it distinguishes reasoning-critical disagreements (positive or negative alignment) from stylistic ones (near-zero alignment) at each token position, without requiring any training.

\begin{figure}[t]
\centering
\resizebox{\textwidth}{!}{%
\input{figures/figure_algo}%
}
\caption{Computing the gradient alignment score at a branching node $u$. (1)~Student rollouts yield empirical $\hat{P}_{\mathrm{succ}}^k$ per branch and the ideal gradient. (2)~A teacher forward pass produces the distillation gradient. (3)~Their cosine similarity measures whether the teacher pushes toward success ($>0$) or against it ($<0$).}
\label{fig:algo-overview}
\end{figure}

\subsection{Computing the score at scale}
\label{sec:computation}

The alignment score (Equation~\ref{eq:alignment}) requires reliable estimates of $P_{\mathrm{succ}}^k$ at each branching node (Figure~\ref{fig:algo-overview} summarizes the three-step computation).
Na\"ively, this would require thousands of rollouts through every possible next token at every node, clearly infeasible for sequences of hundreds of tokens with vocabularies of 150K.

The core challenge is sparsity: given $G$ initial rollouts, most tokens at most nodes receive zero visits, and deep branching points may have only 1--2 rollouts passing through them.
To address this, we generate \emph{targeted rollouts}: given a node $u$ and a token $k$ that needs more visits, we construct a prefix (prompt $+$ path to $u$ $+$ token $k$) and sample completions from the student to the end of the response.
Each targeted rollout enriches not only the target node but all ancestors and descendants along its path, so a single rollout launched at depth $d$ simultaneously improves $P_{\mathrm{succ}}$ estimates at every node it passes through.
This cascading effect means the total budget required grows sublinearly with sequence length.

\paragraph{Exponential depth windows.}
Rather than allocating rollouts uniformly across the sequence, we partition the generation into exponentially growing depth windows (e.g., tokens 1--50, 51--150, 151--350, \ldots).
Within each window, we allocate a fixed budget of $k$ tokens ranked by GKD gradient magnitude and $r$ tokens ranked by student--teacher probability difference, prioritizing tokens where the teacher disagrees most strongly.
Early windows are small and densely sampled (where branching is frequent); later windows are larger and more sparsely sampled (where reasoning chains have committed to a direction).
This mirrors the natural structure of generation trees: branching diversity decreases with depth as trajectories converge.

\paragraph{Budget and scalability.}
We target tokens where $P_\theta^k > \tau$ or $P_{\mathrm{te}}^k > \tau$ ($\tau = 0.02$), i.e., those that contribute meaningfully to the gradient, and enrich each to $N_{\min} = 100$ visits.
Nodes with $\geq N_{\mathrm{sig}} = 20$ visits per child are retained for the alignment computation; for longer traces (AIME) where estimates are noisier, we use $N_{\mathrm{sig}} = 40$.
The total compute scales with the user-chosen budget (number of windows $\times$ per-window budget) rather than with sequence length, making the method applicable to traces ranging from $\sim$200 tokens (BoolQ) to $\sim$30K tokens (AIME) without modification.
In practice, each question requires $\sim$45K--200K targeted rollouts depending on trace length.

\paragraph{Teacher-independent tree sharing.}
A key efficiency insight is that the generation tree and $P_{\mathrm{succ}}^k$ estimates are \emph{teacher-independent}: they depend only on the student's rollouts and outcomes.
We share a single enriched tree across all 8 teacher configurations: rollout generation is done once (Phase~1), and each teacher requires only one additional forward pass to compute its gradient and alignment score (Phase~2).
This reduces total compute by $\sim$7$\times$ compared to independent runs.
Details on rollout prioritization are in Appendix~\ref{app:computation}.

%% file: figures/figure_algo.tex

\definecolor{fig@darkgold}{RGB}{165,125,0}
\definecolor{fig@nodebg}{RGB}{245,245,245}
\definecolor{fig@prefixbg}{RGB}{230,235,245}
\definecolor{fig@branchbg}{RGB}{255,240,210}
\definecolor{fig@panelbg}{RGB}{248,250,254}

\tikzset{
  figA/tok/.style={
    rectangle, rounded corners=2pt,
    minimum height=0.55cm, minimum width=0.60cm,
    font=\small\ttfamily, inner sep=3pt,
    draw=gray!50, fill=fig@nodebg
  },
  figA/tok_prefix/.style={figA/tok, fill=fig@prefixbg, draw=gray!40},
  figA/tok_correct/.style={figA/tok, fill=correctgreen!12, draw=correctgreen!50,
    font=\small\ttfamily\bfseries},
  figA/tok_wrong/.style={figA/tok, fill=misalignred!12, draw=misalignred!45,
    font=\small\ttfamily\bfseries},
  figA/tok_leaf_c/.style={
    rectangle, rounded corners=1.5pt,
    minimum height=0.35cm, minimum width=0.45cm,
    font=\scriptsize\ttfamily\bfseries, inner sep=1.5pt,
    fill=correctgreen!12, draw=correctgreen!50
  },
  figA/tok_leaf_w/.style={
    rectangle, rounded corners=1.5pt,
    minimum height=0.35cm, minimum width=0.45cm,
    font=\scriptsize\ttfamily\bfseries, inner sep=1.5pt,
    fill=misalignred!12, draw=misalignred!45
  },
  figA/model_box/.style={
    rectangle, rounded corners=3pt, draw=alignblue!55,
    fill=alignblue!8, inner sep=6pt, font=\small
  },
  figA/ideal_arr/.style={
    -{Stealth[length=6pt,width=5pt]},
    fig@darkgold, line width=2.5pt, dashed, dash pattern=on 4pt off 2pt
  },
  figA/distill_pos/.style={
    -{Stealth[length=6pt,width=5pt]},
    correctgreen!80!black, line width=2.5pt
  },
  figA/tree_edge/.style={-{Stealth[length=4pt]}, gray!50, line width=0.8pt},
  figA/tree_edge_good/.style={-{Stealth[length=4pt]}, correctgreen!50, line width=1.0pt},
  figA/tree_edge_bad/.style={-{Stealth[length=4pt]}, wrongred!45, line width=1.0pt},
  figA/fan_edge/.style={-{Stealth[length=3pt,width=2pt]}, gray!30, line width=0.5pt},
  figA/flow_arr/.style={
    -{Stealth[length=6pt,width=4.5pt]},
    black!40, line width=1.5pt
  },
}

\begin{tikzpicture}[>=Stealth]


\foreach \xl/\xr in {0/10.0, 10.7/16.5, 17.2/22.5} {
  \draw[draw=gray!35, rounded corners=5pt, fill=fig@panelbg, line width=0.6pt]
    (\xl, -5.5) rectangle (\xr, 0.7);
}


\node[font=\small\bfseries, text=black!80] at (5.0, 0.35)
    {\textcircled{\raisebox{-0.9pt}{\small 1}} Estimate $\hat{P}_{\mathrm{succ}}$ and $\mathbf{g}_{\mathrm{ideal}}$};

\node[figA/tok_prefix] (p1_t0) at (1.4,  -0.30) {\texttt{<think>}};
\node[figA/tok_prefix] (p1_t1) at (2.7,  -0.30) {\texttt{Okay}};
\node[figA/tok_prefix] (p1_t2) at (3.5,  -0.30) {\texttt{,}};
\node[figA/tok_prefix] (p1_t3) at (4.1,  -0.30) {\texttt{so}};
\node[font=\small, text=gray!45] (p1_dots) at (4.7, -0.30) {$\cdots$};
\node[figA/tok_prefix, draw=black!60, line width=0.8pt] (p1_branch) at (5.6, -0.30) {\texttt{= 15}};

\node[figA/tok] (p1_v1) at (1.50, -1.40) {\texttt{so}};
\node[figA/tok] (p1_v2) at (5.00, -1.40) {\texttt{but}};
\node[figA/tok] (p1_v3) at (8.50, -1.40) {\texttt{, wait}};

\draw[figA/tree_edge] (p1_branch.south) -- (p1_v1.north);
\draw[figA/tree_edge] (p1_branch.south) -- (p1_v2.north);
\draw[figA/tree_edge] (p1_branch.south) -- (p1_v3.north);

\node[font=\small, text=gray!45] (p1_d1) at (1.50, -2.20) {$\cdots$};
\node[font=\small, text=gray!45] (p1_d2) at (5.00, -2.20) {$\cdots$};
\node[font=\small, text=gray!45] (p1_d3) at (8.50, -2.20) {$\cdots$};

\draw[figA/tree_edge] (p1_v1.south) -- (p1_d1.north);
\draw[figA/tree_edge] (p1_v2.south) -- (p1_d2.north);
\draw[figA/tree_edge] (p1_v3.south) -- (p1_d3.north);

\node[font=\scriptsize\bfseries, text=correctgreen!80!black] (p1_l1a) at (0.40, -2.85) {$\checkmark$};
\node[font=\scriptsize\bfseries, text=correctgreen!80!black] (p1_l1b) at (0.85, -2.85) {$\checkmark$};
\node[font=\scriptsize\bfseries, text=correctgreen!80!black] (p1_l1c) at (1.30, -2.85) {$\checkmark$};
\node[font=\scriptsize, text=gray!45] at (1.65, -2.85) {$\cdots$};
\node[font=\scriptsize\bfseries, text=correctgreen!80!black] (p1_l1d) at (2.00, -2.85) {$\checkmark$};
\node[font=\scriptsize\bfseries, text=wrongred!80] (p1_l1e) at (2.45, -2.85) {$\times$};

\draw[figA/fan_edge] (p1_d1) -- (p1_l1a.north);
\draw[figA/fan_edge] (p1_d1) -- (p1_l1b.north);
\draw[figA/fan_edge] (p1_d1) -- (p1_l1c.north);
\draw[figA/fan_edge] (p1_d1) -- (p1_l1d.north);
\draw[figA/fan_edge] (p1_d1) -- (p1_l1e.north);

\node[font=\scriptsize\bfseries, text=correctgreen!80!black] at (1.50, -3.35)
    {$\hat{P}_{\mathrm{succ}}\!=\!0.75$};

\node[font=\scriptsize\bfseries, text=correctgreen!80!black] (p1_l2a) at (3.90, -2.85) {$\checkmark$};
\node[font=\scriptsize\bfseries, text=wrongred!80] (p1_l2b) at (4.35, -2.85) {$\times$};
\node[font=\scriptsize, text=gray!45] at (4.70, -2.85) {$\cdots$};
\node[font=\scriptsize\bfseries, text=correctgreen!80!black] (p1_l2c) at (5.10, -2.85) {$\checkmark$};
\node[font=\scriptsize\bfseries, text=wrongred!80] (p1_l2d) at (5.55, -2.85) {$\times$};
\node[font=\scriptsize\bfseries, text=wrongred!80] (p1_l2e) at (6.00, -2.85) {$\times$};

\draw[figA/fan_edge] (p1_d2) -- (p1_l2a.north);
\draw[figA/fan_edge] (p1_d2) -- (p1_l2b.north);
\draw[figA/fan_edge] (p1_d2) -- (p1_l2c.north);
\draw[figA/fan_edge] (p1_d2) -- (p1_l2d.north);
\draw[figA/fan_edge] (p1_d2) -- (p1_l2e.north);

\node[font=\scriptsize, text=neutralgray] at (5.00, -3.35)
    {$\hat{P}_{\mathrm{succ}}\!=\!0.40$};

\node[font=\scriptsize\bfseries, text=wrongred!80] (p1_l3a) at (7.40, -2.85) {$\times$};
\node[font=\scriptsize\bfseries, text=wrongred!80] (p1_l3b) at (7.85, -2.85) {$\times$};
\node[font=\scriptsize\bfseries, text=correctgreen!80!black] (p1_l3c) at (8.30, -2.85) {$\checkmark$};
\node[font=\scriptsize, text=gray!45] at (8.65, -2.85) {$\cdots$};
\node[font=\scriptsize\bfseries, text=wrongred!80] (p1_l3d) at (9.00, -2.85) {$\times$};
\node[font=\scriptsize\bfseries, text=wrongred!80] (p1_l3e) at (9.45, -2.85) {$\times$};

\draw[figA/fan_edge] (p1_d3) -- (p1_l3a.north);
\draw[figA/fan_edge] (p1_d3) -- (p1_l3b.north);
\draw[figA/fan_edge] (p1_d3) -- (p1_l3c.north);
\draw[figA/fan_edge] (p1_d3) -- (p1_l3d.north);
\draw[figA/fan_edge] (p1_d3) -- (p1_l3e.north);

\node[font=\scriptsize\bfseries, text=wrongred!80] at (8.50, -3.35)
    {$\hat{P}_{\mathrm{succ}}\!=\!0.11$};

\node[draw=gray!25, rounded corners=2pt, fill=white, inner sep=4pt,
      font=\small, align=center] at (2.30, -4.50)
    {$\hat{P}_{\mathrm{succ}}^k = S_u^k \!\big/\! N_u^k$};

\node[draw=fig@darkgold!40, rounded corners=2pt, fill=fig@darkgold!7, inner sep=4pt,
      font=\small, align=center] at (7.20, -4.50)
    {$\mathbf{g}_{\mathrm{ideal}}^j \approx P_\theta^j\!\left(\hat{P}_{\mathrm{succ}}^j - \hat{\bar{P}}_{\mathrm{succ}}\right)$};


\node[font=\small\bfseries, text=black!80] at (13.60, 0.35)
    {\textcircled{\raisebox{-0.9pt}{\small 2}} Teacher forward pass};

\node[figA/model_box] (p3_te) at (13.60, -0.70) {teacher $\pi_{\mathrm{te}}$};

\node[font=\small, below=0.55cm of p3_te, align=center] (p3_out)
    {$\log P_{\mathrm{te}}^k$};
\draw[-{Stealth[length=3pt]}, gray!40] (p3_te) -- (p3_out);

\node[draw=gray!30, rounded corners=2pt, fill=white, inner sep=4pt,
      font=\small, below=0.45cm of p3_out, align=center] (p3_lr)
    {$\ell_k \;=\; \log P_\theta^k - \log P_{\mathrm{te}}^k$};

\node[font=\scriptsize\itshape, text=gray!85, below=0.40cm of p3_lr, align=center]
    {GKD: full vocab\\Single-sample: sampled token only};

\node[draw=alignblue!30, rounded corners=2pt, fill=alignblue!5, inner sep=4pt,
      font=\small, below=1.10cm of p3_lr, align=center] (p3_grad)
    {$\mathbf{g}^{\mathrm{KD}}_j = P_\theta^j\!\left(\ell_j - \bar{\ell}\right)$};


\node[font=\small\bfseries, text=black!80] at (19.85, 0.35)
    {\textcircled{\raisebox{-0.9pt}{\small 3}} Compute $\mathrm{Align}(u)$};

\node[draw=gray!35, rounded corners=3pt, fill=gray!4,
      minimum width=2.8cm, minimum height=2.0cm] (p4_gbox) at (19.85, -1.40) {};
\coordinate (p4_o) at (p4_gbox.center);

\draw[->, gray!25, line width=0.5pt]
    ($(p4_o)+(-1.1,0)$) -- ($(p4_o)+(1.1,0)$);
\draw[->, gray!25, line width=0.5pt]
    ($(p4_o)+(0,-0.80)$) -- ($(p4_o)+(0,0.80)$);

\draw[figA/ideal_arr] (p4_o) -- ++(0.40, 0.62) coordinate (p4_gGRPO);
\node[font=\scriptsize\bfseries, text=fig@darkgold, above=1pt of p4_gGRPO, anchor=south]
    {$\mathbf{g}^{\mathrm{ideal}}$};

\draw[figA/distill_pos] (p4_o) -- ++(0.68, 0.32) coordinate (p4_gD);
\node[font=\scriptsize\bfseries, text=correctgreen!80!black, right=1pt of p4_gD, anchor=west]
    {$\mathbf{g}^{\mathrm{distill}}$};

\draw[gray!50, thin]
    ($(p4_o)+(0.28,0.17)$) arc[start angle=25, end angle=57, radius=0.33cm];
\node[font=\scriptsize, text=gray!60] at ($(p4_o)+(0.38,0.38)$) {$\alpha$};

\node[font=\scriptsize\itshape, text=gray!55, above=2pt of p4_gbox.north]
    {$\nabla_\theta L$ at $u$};

\node[font=\small, below=0.4cm of p4_gbox.south, anchor=north, align=center]
    (p4_formula)
    {$\mathrm{Align}(u) = \cos\alpha$};

\node[font=\scriptsize, below=4pt of p4_formula, align=center]
    {\textcolor{correctgreen!80!black}{$>0$: KD aligned with ideal}\\[1pt]
     \textcolor{misalignred}{$<0$: in conflict}};


\draw[figA/flow_arr] (10.03, -2.0) -- (10.67, -2.0);
\draw[figA/flow_arr] (16.43, -2.0) -- (17.17, -2.0);

\end{tikzpicture}

%% file: related_work_short.tex
\section{Related Work}
\label{sec:related}

Knowledge distillation~\citep{hinton2015distilling} trains a student on teacher soft distributions; sequence-level variants~\citep{kim2016sequence} generate teacher outputs for training but suffer from exposure bias~\citep{bengio2015scheduled}.
On-policy distillation (OPD) addresses this by supervising the student on its own rollouts: MiniLLM~\citep{gu2024minillm} uses reverse KL via policy gradient, GKD~\citep{agarwal2024gkd} interpolates between on- and off-policy data, and \citet{yang2026beyond} show the teacher's log-ratio acts as an implicit dense reward.
OPD is now standard in industry pipelines~\citep{qwen3, xiao2026mimo, zeng2026glm5}, and has been extended to self-distillation settings where a single model serves as its own teacher under privileged conditioning~\citep{snell2022learning, hubotter2026sdpo, zhao2026opsd, ye2026onpolicy, shenfeld2026continual, penaloza2026privileged}.
Our work does not propose a new distillation algorithm but provides a diagnostic that measures, at each token, whether the teacher's signal agrees with the reward objective.

Despite OPD's growing adoption, when and why it fails remains poorly understood.
Two concurrent works investigate this question from complementary angles.
\citet{li2026rethinking} show that OPD requires thinking-pattern consistency and genuinely new knowledge from the teacher, with success driven by progressive alignment on high-probability overlap tokens (97--99\% of mass); they also reveal that reward quality degrades with trajectory depth.
\citet{kim2026selfdistill} trace self-distillation degradation to the suppression of epistemic verbalization (the model's expression of uncertainty), showing that richer conditioning contexts suppress uncertainty tokens and harm OOD generalization when task coverage is broad.
More broadly, distillation can hurt with overly capable teachers~\citep{cho2019efficacy, mirzadeh2020teacher, busbridge2025scaling}, and small models struggle to learn from strong reasoners~\citep{li2025small}.
Our gradient alignment score provides a mechanistic explanation for these phenomena: it directly quantifies, at each token position, whether the teacher's signal is exploitable by the student, regardless of the teacher's aggregate performance.

On the reward side, GRPO~\citep{shao2024deepseekmath, deepseek2025r1, liu2025understanding} and DAPO~\citep{yu2025dapo} train reasoning models with sparse outcome rewards, while process reward models~\citep{lightman2024prm, uesato2022solving} provide step-level feedback but require separate annotation.
Our gradient decomposition unifies these perspectives by showing that reward and distillation objectives share the same local structure ($P_\theta^j(f_j - \bar{f})$), enabling direct offline comparison at token granularity without training or additional models.

%% file: aime_results.tex

\subsection{Overview}

To test whether our findings extend to mathematical reasoning (where thinking traces are substantially longer, $\sim$5K--30K tokens, and reasoning steps more complex), we analyze four AIME 2025 questions: two with Qwen3-0.6B (Q0, pass rate 87.5\%; Q3, pass rate 9.4\%) and two with Qwen3-1.7B (Q7, pass rate 34.4\%; Q28, pass rate 56.3\%).
Each question uses the same 8 teacher configurations as BoolQ/MMLU.
We use sig${}=40$ (requiring $\geq$40 visits per child for significance) given the longer chains and sparser branching.

\begin{figure}[t]
\centering
\includegraphics[width=\textwidth]{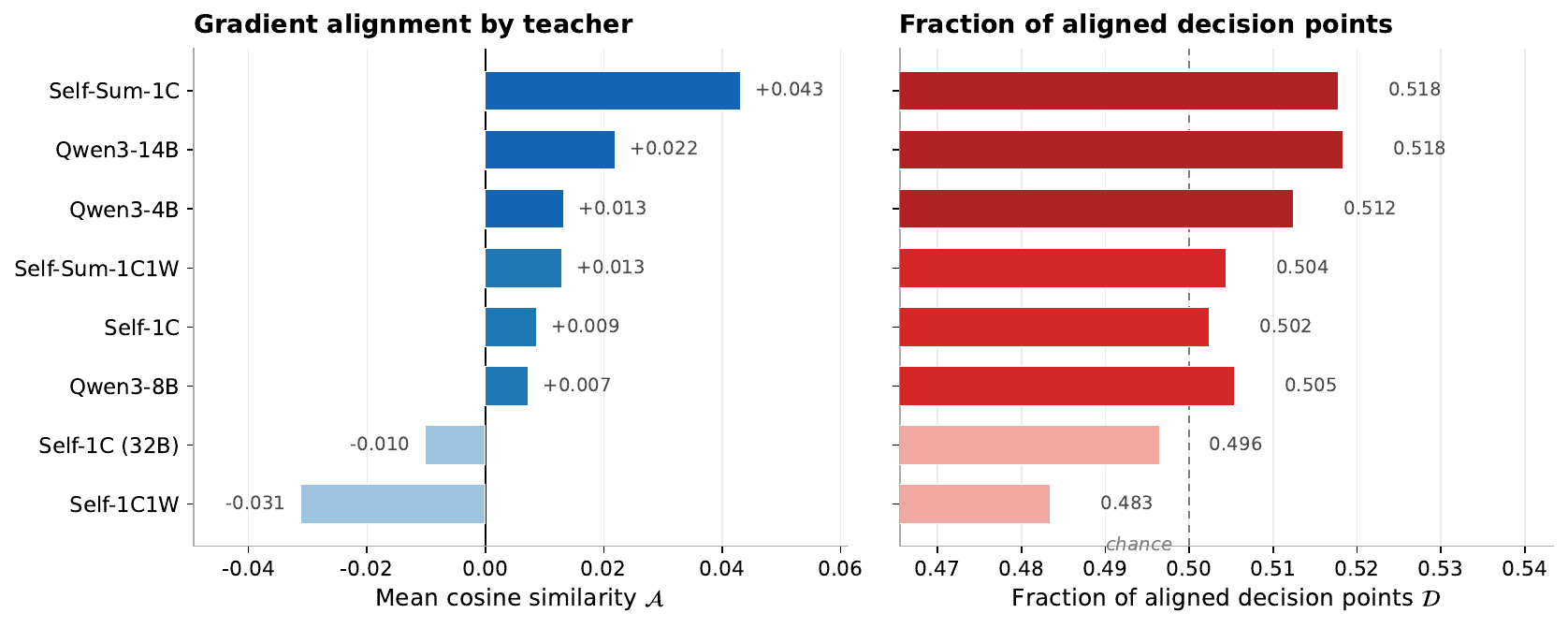}
\captionof{figure}{AIME Q0, Qwen3-0.6B (pass rate 87.5\%)}
\vspace{6pt}
\includegraphics[width=\textwidth]{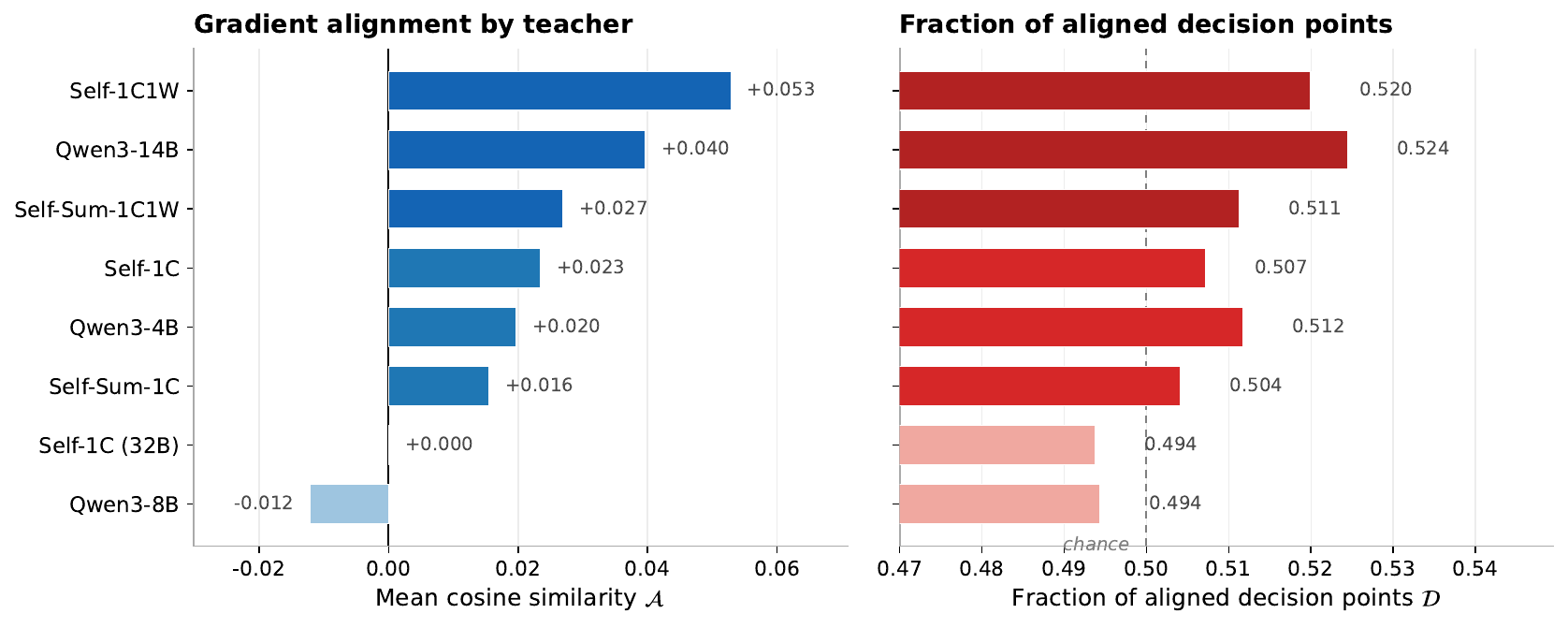}
\captionof{figure}{AIME Q28, Qwen3-1.7B (pass rate 56.3\%)}
\caption{Teacher ranking by gradient alignment on two AIME 2025 questions. Both external teachers and self-distillation variants achieve positive alignment, with \textsc{Self-Sum-1C} and \textsc{Self-1C1W} leading.}
\label{fig:aime-teachers}
\end{figure}

\paragraph{Teacher rankings on AIME.}
Figure~\ref{fig:aime-teachers} shows teacher alignment on two representative questions.
Both external and self-distillation teachers achieve positive alignment across all four questions.
\textsc{Self-Sum-1C} (summarized correct demonstration) is the strongest teacher on Q0 ($+0.043$) and Q7 ($+0.055$), while the raw demo variants lead on Q3 (\textsc{Self-1C1W}: $+0.042$, \textsc{Self-1C}: $+0.040$) and Q28 (\textsc{Self-1C1W}: $+0.053$).
Notably, \textsc{Self-1C1W} (which includes a wrong demonstration) is the best teacher on the two harder questions (Q3 and Q28), contrasting sharply with BoolQ/MMLU where it consistently underperforms.

\paragraph{Correct vs.\ incorrect on AIME.}
Among teachers with positive alignment, the incorrect $>$ correct pattern holds on all four questions when focusing on high-stakes decision points.
On Q0, all positive teachers show higher alignment on incorrect paths (best teacher \textsc{Self-Sum-1C}: incorrect $+0.097$ vs.\ correct $-0.011$, $\Delta = -0.108$).
On Q3, the top-ranked teachers show the same pattern (\textsc{Self-1C1W}: incorrect $+0.056$ vs.\ correct $+0.027$, $\Delta = -0.029$).
On Q28, 7 of 8 positive teachers show incorrect $>$ correct (\textsc{Self-1C1W}: incorrect $+0.097$ vs.\ correct $+0.008$, $\Delta = -0.089$).
On Q7, the pattern emerges clearly at consequential nodes (best teacher \textsc{Self-Sum-1C}: incorrect $+0.093$ vs.\ correct $+0.017$, $\Delta = -0.076$ at high success-rate-range nodes), though the gap is smaller at low-stakes nodes where the signal is noisier.
This confirms that even on long mathematical reasoning chains, the teacher's gradient is most useful on the student's failing rollouts.

\paragraph{What differs from BoolQ/MMLU.}
While the correct/incorrect finding transfers, teacher-choice conclusions do not, further supporting the comprehensibility hypothesis.
On hard AIME math (Q3), \textsc{Self-1C1W} (which includes a wrong demonstration alongside the correct one) is the \emph{best} teacher ($+0.042$), directly contradicting BoolQ/MMLU where wrong demonstrations consistently hurt.
On hard math, seeing a common mistake may provide useful contrastive signal (``avoid this error'') that the student can comprehend and act on, whereas on simpler tasks the wrong solution is merely distracting noise.
Meanwhile, \textsc{Self-Sum-1C} (summarized) loses to raw \textsc{Self-1C} on Q3: the 0.6B student cannot decipher a compressed summary of a hard mathematical argument and needs the full step-by-step trace.
On easier/medium questions (Q0, Q7), summarized contexts perform well because the reasoning is simple enough to compress without losing comprehensibility.

Additionally, teacher rankings shift with filter stringency: on Q7, external teachers lead at lenient filters but \textsc{Self-Sum-1C} dominates at strict filters that focus on high-stakes nodes.
These observations reinforce that no universal distillation recipe exists: the optimal teacher configuration depends on the task, the question difficulty, the student's capacity to comprehend the context, and which decision points one prioritizes.

%% file: prompts.tex
\begin{figure}[h!]
\centering
\begin{tcolorbox}[
  enhanced,
  width=0.93\linewidth,
  colback=white,
  colframe=promptborder,
  boxrule=0.5pt,
  arc=4pt,
  left=1pt, right=1pt, top=3pt, bottom=3pt,
  fontupper=\small\ttfamily,
]
\begin{tcolorbox}[
  enhanced, arc=6pt,
  colback=systembg, colframe=systemframe!40,
  boxrule=0.4pt,
  left=4pt, right=4pt, top=2pt, bottom=2pt,
  width=\linewidth,
]
{\sffamily\scriptsize\bfseries\color{systemframe} SYSTEM}\par\smallskip
You are a helpful assistant.
\end{tcolorbox}
\vspace{2pt}
\begin{tcolorbox}[
  enhanced, arc=6pt,
  colback=userbg, colframe=userframe!40,
  boxrule=0.4pt,
  left=4pt, right=4pt, top=2pt, bottom=2pt,
  width=\linewidth,
]
{\sffamily\scriptsize\bfseries\color{userframe} USER}\par\smallskip
Answer the following multiple choice question. The last line of your response should be of the following format: 'Answer: \$LETTER' (without quotes) where LETTER is one of ABCD. Think step by step before answering.\par\smallskip
\{question\}\par\smallskip
A) \{choice\_A\}\\
B) \{choice\_B\}\\
C) \{choice\_C\}\\
D) \{choice\_D\}
\end{tcolorbox}
\end{tcolorbox}
\caption{Baseline prompt (no context). Shown for MMLU (multiple-choice).}
\label{fig:prompt_baseline}
\end{figure}

\begin{figure}[h!]
\centering
\begin{tcolorbox}[
  enhanced,
  width=0.93\linewidth,
  colback=white,
  colframe=promptborder,
  boxrule=0.5pt,
  arc=4pt,
  left=1pt, right=1pt, top=3pt, bottom=3pt,
  fontupper=\small\ttfamily,
]
\begin{tcolorbox}[
  enhanced, arc=6pt,
  colback=systembg, colframe=systemframe!40,
  boxrule=0.4pt,
  left=4pt, right=4pt, top=2pt, bottom=2pt,
  width=\linewidth,
]
{\sffamily\scriptsize\bfseries\color{systemframe} SYSTEM}\par\smallskip
You are a helpful assistant.
\end{tcolorbox}
\vspace{2pt}
\begin{tcolorbox}[
  enhanced, arc=6pt,
  colback=userbg, colframe=userframe!40,
  boxrule=0.4pt,
  left=4pt, right=4pt, top=2pt, bottom=2pt,
  width=\linewidth,
]
{\sffamily\scriptsize\bfseries\color{userframe} USER}\par\smallskip
\{question with instruction + choices\}\par\smallskip
\begin{tcolorbox}[
  enhanced, arc=4pt,
  colback=demoframe!8, colframe=demoframe!30,
  boxrule=0.3pt,
  left=4pt, right=4pt, top=2pt, bottom=2pt,
]
{\sffamily\scriptsize\color{demoframe!80!black}\bfseries Demonstration}\par\smallskip
This is a correct response to the question:\par
\textrm{"""}\par
\{correct\_response\_1\}\par
\textrm{"""}
\end{tcolorbox}
\smallskip
Now answer with a response of your own, including the thinking process:
\end{tcolorbox}
\end{tcolorbox}
\caption{\textsc{Self-1C} prompt. A single correct demonstration is prepended. For \textsc{Self-1C\,(32B)} the demonstration is generated by Qwen3-32B instead; the prompt format is identical.}
\label{fig:prompt_demo1c}
\end{figure}

\begin{figure}[h!]
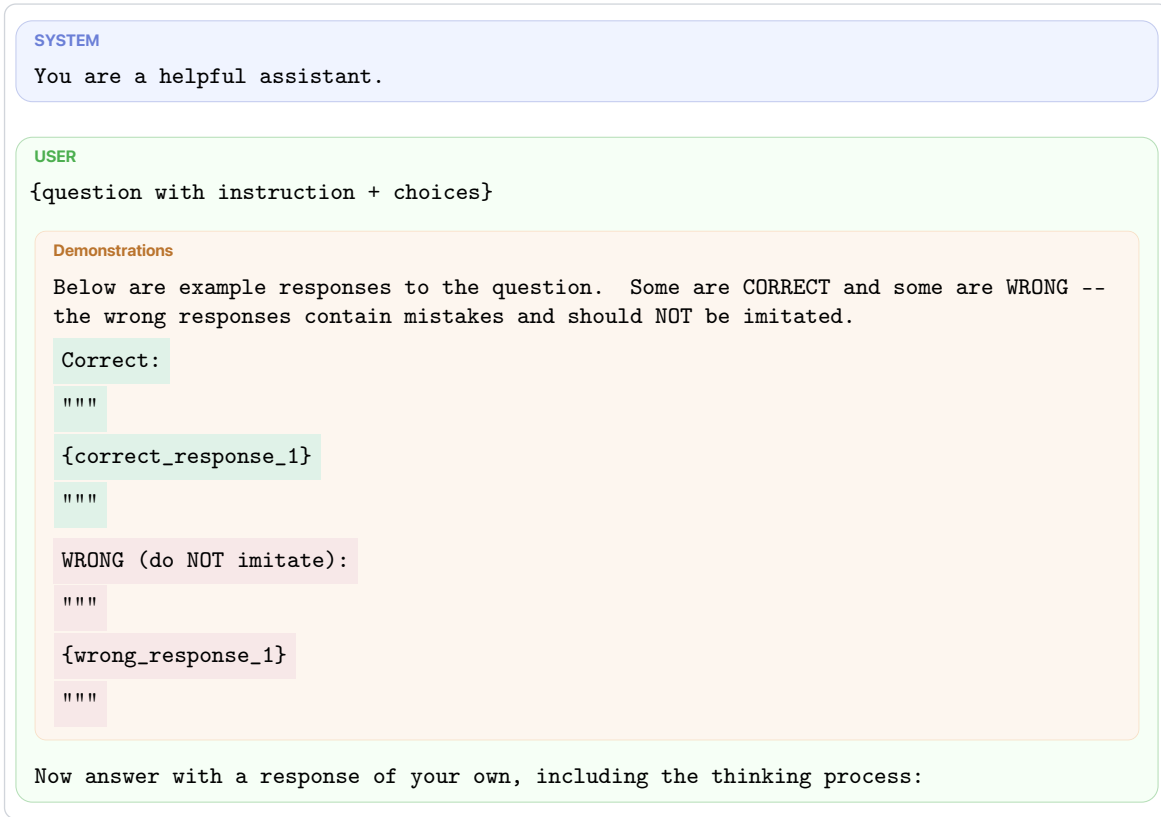

\centering
\begin{tcolorbox}[
  enhanced,
  width=0.93\linewidth,
  colback=white,
  colframe=promptborder,
  boxrule=0.5pt,
  arc=4pt,
  left=1pt, right=1pt, top=3pt, bottom=3pt,
  fontupper=\small\ttfamily,
]
\begin{tcolorbox}[
  enhanced, arc=6pt,
  colback=systembg, colframe=systemframe!40,
  boxrule=0.4pt,
  left=4pt, right=4pt, top=2pt, bottom=2pt,
  width=\linewidth,
]
{\sffamily\scriptsize\bfseries\color{systemframe} SYSTEM}\par\smallskip
You are a helpful assistant.
\end{tcolorbox}
\vspace{2pt}
\begin{tcolorbox}[
  enhanced, arc=6pt,
  colback=userbg, colframe=userframe!40,
  boxrule=0.4pt,
  left=4pt, right=4pt, top=2pt, bottom=2pt,
  width=\linewidth,
]
{\sffamily\scriptsize\bfseries\color{userframe} USER}\par\smallskip
\{question with instruction + choices\}\par\smallskip
\begin{tcolorbox}[
  enhanced, arc=4pt,
  colback=demoframe!8, colframe=demoframe!30,
  boxrule=0.3pt,
  left=4pt, right=4pt, top=2pt, bottom=2pt,
]
{\sffamily\scriptsize\color{demoframe!80!black}\bfseries Demonstrations}\par\smallskip
Below are example responses to the question. Some are CORRECT and some are WRONG --- the wrong responses contain mistakes and should NOT be imitated.\par\smallskip
\colorbox{correctgreen!12}{\strut Correct:}\\
\colorbox{correctgreen!12}{\strut \textrm{"""}}\\
\colorbox{correctgreen!12}{\strut \{correct\_response\_1\}}\\
\colorbox{correctgreen!12}{\strut \textrm{"""}}\par\smallskip
\colorbox{wrongred!10}{\strut WRONG (do NOT imitate):}\\
\colorbox{wrongred!10}{\strut \textrm{"""}}\\
\colorbox{wrongred!10}{\strut \{wrong\_response\_1\}}\\
\colorbox{wrongred!10}{\strut \textrm{"""}}
\end{tcolorbox}
\smallskip
Now answer with a response of your own, including the thinking process:
\end{tcolorbox}
\end{tcolorbox}
\caption{\textsc{Self-1C1W} prompt. One correct and one wrong demonstration with warning header.}
\label{fig:prompt_demo1c1w}
\end{figure}

\begin{figure}[h!]
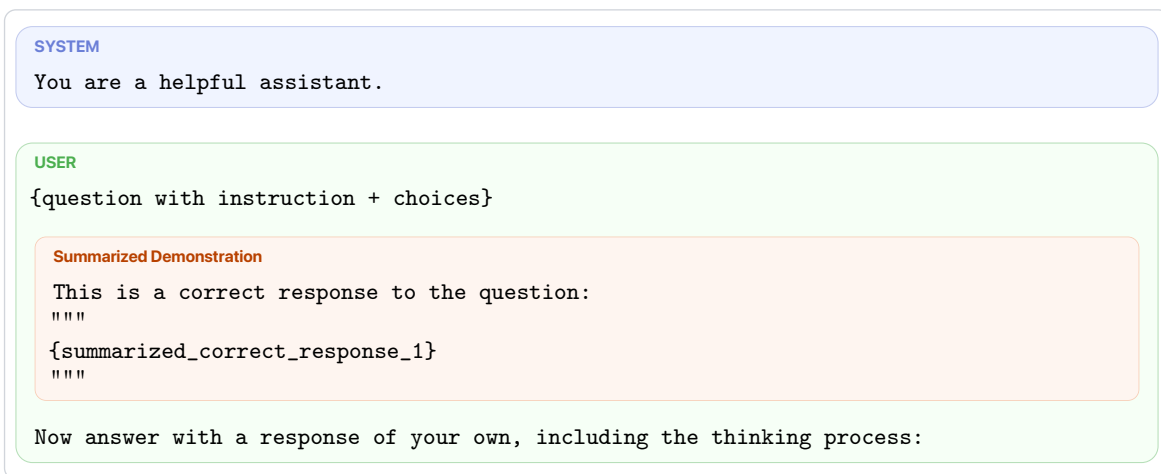

\centering
\begin{tcolorbox}[
  enhanced,
  width=0.93\linewidth,
  colback=white,
  colframe=promptborder,
  boxrule=0.5pt,
  arc=4pt,
  left=1pt, right=1pt, top=3pt, bottom=3pt,
  fontupper=\small\ttfamily,
]
\begin{tcolorbox}[
  enhanced, arc=6pt,
  colback=systembg, colframe=systemframe!40,
  boxrule=0.4pt,
  left=4pt, right=4pt, top=2pt, bottom=2pt,
  width=\linewidth,
]
{\sffamily\scriptsize\bfseries\color{systemframe} SYSTEM}\par\smallskip
You are a helpful assistant.
\end{tcolorbox}
\vspace{2pt}
\begin{tcolorbox}[
  enhanced, arc=6pt,
  colback=userbg, colframe=userframe!40,
  boxrule=0.4pt,
  left=4pt, right=4pt, top=2pt, bottom=2pt,
  width=\linewidth,
]
{\sffamily\scriptsize\bfseries\color{userframe} USER}\par\smallskip
\{question with instruction + choices\}\par\smallskip
\begin{tcolorbox}[
  enhanced, arc=4pt,
  colback=summcolor!6, colframe=summcolor!30,
  boxrule=0.3pt,
  left=4pt, right=4pt, top=2pt, bottom=2pt,
]
{\sffamily\scriptsize\color{summcolor!80!black}\bfseries Summarized Demonstration}\par\smallskip
This is a correct response to the question:\par
\textrm{"""}\par
\{summarized\_correct\_response\_1\}\par
\textrm{"""}
\end{tcolorbox}
\smallskip
Now answer with a response of your own, including the thinking process:
\end{tcolorbox}
\end{tcolorbox}
\caption{\textsc{Self-Sum-1C} prompt. The demonstration is first condensed by Qwen3-32B.}
\label{fig:prompt_summ1c}
\end{figure}

\vspace{0.3cm}

%% file: tables/screening_details.tex
\begin{table}[t]
\centering
\caption{Number of questions with at least one correct demonstration available among randomly selected questions for screening, and the intersection used for analysis.}
\label{tab:context_filter}
\small
\begin{tabular}{llrrr}
\toprule
\textbf{Model} & \textbf{Benchmark} & \textbf{Self} & \textbf{32B} & \textbf{Both} \\
\midrule
Qwen3-0.6B & MMLU    & $381$ & $459$ & $364$ \\
Qwen3-0.6B & BoolQ   & $473$ & $482$ & $458$ \\
\midrule
Qwen3-1.7B & MMLU    & $357$ & $459$ & $350$ \\
Qwen3-1.7B & BoolQ   & $474$ & $482$ & $463$ \\
\bottomrule
\end{tabular}
\end{table}

%% file: tables/screening_results.tex
\begin{table*}[t]
\centering
\caption{Screening results. $\Delta$ is the absolute improvement over baseline.}
\label{tab:screening_mmlu_boolq}
\small
\setlength{\tabcolsep}{4pt}
\begin{tabular}{l cc cc cc cc}
\toprule
& \multicolumn{4}{c}{\textbf{MMLU}} & \multicolumn{4}{c}{\textbf{BoolQ}} \\
\cmidrule(lr){2-5} \cmidrule(lr){6-9}
& \multicolumn{2}{c}{Qwen3-0.6B} & \multicolumn{2}{c}{Qwen3-1.7B} & \multicolumn{2}{c}{Qwen3-0.6B} & \multicolumn{2}{c}{Qwen3-1.7B} \\
\cmidrule(lr){2-3} \cmidrule(lr){4-5} \cmidrule(lr){6-7} \cmidrule(lr){8-9}
\textbf{Context Variant} & Acc & $\Delta$ & Acc & $\Delta$ & Acc & $\Delta$ & Acc & $\Delta$ \\
\midrule
Baseline             & $63.8$ & ---      & $85.2$ & ---      & $78.5$ & ---      & $88.9$ & ---      \\
\textsc{Self-1C}     & $98.6$ & $+35.3$  & $99.7$ & $+14.5$  & $98.5$ & $+19.9$  & $98.3$ & $+10.0$  \\
\textsc{Self-1C1W}   & $68.2$ & $+4.7$   & $94.0$ & $+8.8$   & $77.6$ & $-0.9$   & $87.7$ & $-0.8$   \\
\textsc{Self-3C}     & $99.4$ & $+36.1$  & $99.8$ & $+14.6$  & $99.3$ & $+20.7$  & $99.3$ & $+11.0$  \\
\textsc{Self-Sum-1C} & $96.4$ & $+32.7$  & $99.3$ & $+14.1$  & $97.4$ & $+18.8$  & $98.2$ & $+9.5$   \\
\textsc{Self-Sum-1C1W} & $91.5$ & $+27.8$  & $98.2$ & $+13.0$  & $92.5$ & $+14.0$  & $94.5$ & $+5.8$   \\
\textsc{Self-1C\,(32B)} & $98.8$ & $+35.5$  & $99.7$ & $+14.5$  & $98.4$ & $+19.9$  & $98.7$ & $+10.4$  \\
\bottomrule
\end{tabular}
\end{table*}

%% file: tables/screening_breakdown.tex
\begin{table*}[t]
\centering
\caption{Detailed difficulty breakdown for \textbf{Qwen3-0.6B} (left) and \textbf{Qwen3-1.7B} (right). Each cell contains accuracy (\%) and number of questions ($n$).}
\label{tab:breakdown}

\begin{minipage}[t]{0.49\textwidth}
\centering
\resizebox{\linewidth}{!}{%
\begin{tabular}{l rr rr rr rr r}
\toprule
\multicolumn{10}{c}{\textbf{Qwen3-0.6B}} \\
\midrule
& \multicolumn{2}{c}{\textbf{Hard}} & \multicolumn{2}{c}{\textbf{Med.}} & \multicolumn{2}{c}{\textbf{Easy}} & \multicolumn{2}{c}{\textbf{All}} & \\
\cmidrule(lr){2-3} \cmidrule(lr){4-5} \cmidrule(lr){6-7} \cmidrule(lr){8-9}
\textbf{Variant} & Acc & $n$ & Acc & $n$ & Acc & $n$ & Acc & $n$ & $\Delta$ \\
\midrule
\multicolumn{10}{c}{\textit{MMLU}} \\
\midrule
Baseline    & $12.6$ & $50$ & $52.9$ & $168$ & $94.6$ & $143$ & $63.8$ & $361$ & ---   \\
\textsc{Self-1C}     & $97.4$ & $50$ & $98.1$ & $168$ & $99.6$ & $143$ & $98.6$ & $361$ & $+35.3$ \\
\textsc{Self-1C1W}   & $41.3$ & $50$ & $59.4$ & $168$ & $88.0$ & $143$ & $68.2$ & $361$ & $+4.7$  \\
\textsc{Self-3C}     & $98.2$ & $50$ & $99.4$ & $168$ & $100.0$ & $143$ & $99.4$ & $361$ & $+36.1$ \\
\textsc{Self-Sum-1C}     & $90.5$ & $50$ & $95.3$ & $168$ & $99.7$ & $143$ & $96.4$ & $361$ & $+32.7$ \\
\textsc{Self-Sum-1C1W}   & $80.0$ & $50$ & $89.0$ & $168$ & $98.3$ & $143$ & $91.5$ & $361$ & $+27.8$ \\
\textsc{Self-1C\,(32B)} & $96.0$ & $50$ & $99.1$ & $168$ & $99.5$ & $143$ & $98.8$ & $361$ & $+35.5$ \\
\midrule
\multicolumn{10}{c}{\textit{BoolQ}} \\
\midrule
Baseline    & $12.9$ & $29$ & $56.2$ & $130$ & $94.6$ & $299$ & $78.5$ & $458$ & ---   \\
\textsc{Self-1C}     & $92.6$ & $29$ & $96.9$ & $130$ & $99.7$ & $299$ & $98.5$ & $458$ & $+19.9$ \\
\textsc{Self-1C1W}   & $35.2$ & $29$ & $57.1$ & $130$ & $90.7$ & $299$ & $77.6$ & $458$ & $-0.9$ \\
\textsc{Self-3C}     & $96.0$ & $29$ & $98.7$ & $130$ & $99.9$ & $299$ & $99.3$ & $458$ & $+20.7$ \\
\textsc{Self-Sum-1C}     & $84.1$ & $29$ & $94.8$ & $130$ & $99.7$ & $299$ & $97.4$ & $458$ & $+18.8$ \\
\textsc{Self-Sum-1C1W}   & $68.6$ & $29$ & $85.1$ & $130$ & $98.1$ & $299$ & $92.5$ & $458$ & $+14.0$ \\
\textsc{Self-1C\,(32B)} & $94.4$ & $29$ & $96.3$ & $130$ & $99.7$ & $299$ & $98.4$ & $458$ & $+19.9$ \\
\bottomrule
\end{tabular}%
}
\end{minipage}%
\hfill
\begin{minipage}[t]{0.49\textwidth}
\centering
\resizebox{\linewidth}{!}{%
\begin{tabular}{l rr rr rr rr r}
\toprule
\multicolumn{10}{c}{\textbf{Qwen3-1.7B}} \\
\midrule
& \multicolumn{2}{c}{\textbf{Hard}} & \multicolumn{2}{c}{\textbf{Med.}} & \multicolumn{2}{c}{\textbf{Easy}} & \multicolumn{2}{c}{\textbf{All}} & \\
\cmidrule(lr){2-3} \cmidrule(lr){4-5} \cmidrule(lr){6-7} \cmidrule(lr){8-9}
\textbf{Variant} & Acc & $n$ & Acc & $n$ & Acc & $n$ & Acc & $n$ & $\Delta$ \\
\midrule
\multicolumn{10}{c}{\textit{MMLU}} \\
\midrule
Baseline    & $12.5$ & $15$ & $53.1$ & $70$ & $97.8$ & $265$ & $85.2$ & $350$ & ---   \\
\textsc{Self-1C}     & $99.0$ & $15$ & $98.8$ & $70$ & $100.0$ & $265$ & $99.7$ & $350$ & $+14.5$ \\
\textsc{Self-1C1W}   & $71.9$ & $15$ & $83.0$ & $70$ & $98.1$ & $265$ & $94.0$ & $350$ & $+8.8$  \\
\textsc{Self-3C}     & $98.8$ & $15$ & $99.4$ & $70$ & $100.0$ & $265$ & $99.8$ & $350$ & $+14.6$ \\
\textsc{Self-Sum-1C}     & $96.5$ & $15$ & $97.1$ & $70$ & $100.0$ & $265$ & $99.3$ & $350$ & $+14.1$ \\
\textsc{Self-Sum-1C1W}   & $85.0$ & $15$ & $94.3$ & $70$ & $100.0$ & $265$ & $98.2$ & $350$ & $+13.0$ \\
\textsc{Self-1C\,(32B)} & $100.0$ & $15$ & $98.5$ & $70$ & $100.0$ & $265$ & $99.7$ & $350$ & $+14.5$ \\
\midrule
\multicolumn{10}{c}{\textit{BoolQ}} \\
\midrule
Baseline    & $12.9$ & $16$ & $53.2$ & $66$ & $98.3$ & $378$ & $88.9$ & $460$ & ---   \\
\textsc{Self-1C}     & $74.6$ & $16$ & $95.0$ & $66$ & $99.9$ & $378$ & $98.3$ & $460$ & $+10.0$ \\
\textsc{Self-1C1W}   & $31.6$ & $16$ & $58.2$ & $66$ & $95.3$ & $378$ & $87.7$ & $460$ & $-0.8$ \\
\textsc{Self-3C}     & $91.0$ & $16$ & $97.5$ & $66$ & $100.0$ & $378$ & $99.3$ & $460$ & $+11.0$ \\
\textsc{Self-Sum-1C}     & $70.3$ & $16$ & $95.3$ & $66$ & $99.8$ & $378$ & $98.2$ & $460$ & $+9.5$  \\
\textsc{Self-Sum-1C1W}   & $55.7$ & $16$ & $76.9$ & $66$ & $99.2$ & $378$ & $94.5$ & $460$ & $+5.8$  \\
\textsc{Self-1C\,(32B)} & $85.9$ & $16$ & $96.1$ & $66$ & $99.7$ & $378$ & $98.7$ & $460$ & $+10.4$ \\
\bottomrule
\end{tabular}%
}
\end{minipage}

\end{table*}